\documentclass[11pt,a4paper]{article}
\usepackage[utf8]{inputenc}

\usepackage[hyphens]{url}
\usepackage{xurl}
\usepackage{breakurl}

\usepackage{mathrsfs,amsthm,xcolor,verbatim,bbm,amsmath,amsfonts,amssymb,nicefrac,enumitem,hyperref,bm,mathtools,xparse,etoolbox,textcomp}
\usepackage[margin=1in]{geometry}
\usepackage[capitalise]{cleveref} 

\crefname{enumi}{item}{items}

\crefname{equation}{}{}
\crefname{subsection}{Subsection}{Subsections}

\numberwithin{equation}{section}

\hypersetup{
    colorlinks,
    linkcolor={red!50!black},
    citecolor={green},
    urlcolor={blue!80!black}
}

\theoremstyle{plain}
\newtheorem{theorem}{Theorem} [section]
\newtheorem{lemma}[theorem]{Lemma}
\newtheorem{prop}[theorem]{Proposition}
\newtheorem{cor}[theorem]{Corollary}

\newtheorem{setting}[theorem]{Setting}

\theoremstyle{remark}

\theoremstyle{definition}
\newtheorem{definition}[theorem]{Definition}

\DeclareFontEncoding{LS1}{}{}
\DeclareFontSubstitution{LS1}{stix}{m}{n}
\DeclareMathAlphabet{\mathscr}{LS1}{stixscr}{m}{n}

\newcommand{\R}{\mathbb{R}}
\newcommand{\N}{\mathbb{N}}

\newcommand{\dens}{\mathfrak{p}}
\newcommand{\const}{\fD}

\newcommand{\w}[1]{\mathfrak{w}^{#1}}
\renewcommand{\b}[1]{\mathfrak{b}^{#1}}
\renewcommand{\v}[1]{\mathfrak{v}^{#1}}

\renewcommand{\c}[1]{\mathfrak{c}^{#1}}

\newcommand{\smallsum}{\textstyle\sum}

\newcommand{\polyn}{\scrP}
\newcommand{\ratio}{\scrR}

\newcommand{\with}{\curvearrowleft}

\newcommand{\sign}{\mathfrak{s}}

\newcommand{\cB}{\mathcal{B}}

\newcommand{\cG}{\mathcal{G}}

\newcommand{\cI}{\mathcal{I}}
\newcommand{\cJ}{\mathcal{J}}

\newcommand{\cL}{\mathcal{L}}

\newcommand{\cS}{\mathcal{S}}

\newcommand{\bfm}{\mathbf{m}}

\newcommand{\bfD}{\mathbf{D}}

\newcommand{\bfM}{\mathbf{M}}

\newcommand{\scrA}{\mathscr{A}}

\newcommand{\scrC}{\mathscr{C}}

\newcommand{\scrN}{\mathscr{N}}

\newcommand{\scrP}{\mathscr{P}}

\newcommand{\scrR}{\mathscr{R}}

\newcommand{\scra}{\mathscr{a}}
\newcommand{\scrb}{\mathscr{b}}

\newcommand{\fC}{\mathfrak{C}}
\newcommand{\fD}{\mathfrak{D}}

\newcommand{\fG}{\mathfrak{G}}

\newcommand{\fL}{\mathfrak{L}}

\newcommand{\fP}{\mathfrak{P}}

\newcommand{\fa}{\mathfrak{a}}
\newcommand{\fb}{\mathfrak{b}}
\newcommand{\fc}{\mathfrak{c}}
\newcommand{\fd}{\mathfrak{d}}

\newcommand{\fs}{\mathfrak{s}}

\newcommand{\fw}{\mathfrak{w}}

\renewcommand{\emptyset}{\varnothing}

\DeclarePairedDelimiter{\norm}{\lVert}{\rVert}
\DeclarePairedDelimiter{\abs}{\lvert}{\rvert}
\DeclarePairedDelimiter{\rbr}{(}{)}
\DeclarePairedDelimiter{\br}{[}{]}
\DeclarePairedDelimiter{\cu}{\{}{\}}
\DeclarePairedDelimiter{\spro}{\langle}{\rangle}

\newcommand{\Rect}{\mathfrak{R}}

\renewcommand{\d}{ \mathrm{d}}

\newcommand{\qandq}{\quad\text{and}\quad}
\newcommand{\qqandqq}{\qquad\text{and}\qquad}

\newcommand{\indicator}[1]{\mathbbm{1}_{\smash{#1}}}

\newcommand{\realization}[1] {\mathscr{N} ^{ #1  }}

\newcommand{\width}{H}

\ExplSyntaxOn

\bool_new:N \g_noteobserve

\NewDocumentCommand{\nobs}{}{
	\bool_if:nTF { \g_noteobserve } {
		\bool_gset_false:N \g_noteobserve 
		note~
	} {
		\bool_gset_true:N \g_noteobserve 
		observe~
	}
}

\NewDocumentCommand{\Nobs}{}{
	\bool_if:nTF { \g_noteobserve } {
		\bool_gset_false:N \g_noteobserve 
		Note~
	} {
		\bool_gset_true:N \g_noteobserve 
		Observe~
	}
}

\seq_new:N \g_cflist_loaded
\seq_new:N \g_cflist_pending

\NewDocumentCommand{\cfadd}{ m }
{
	\seq_if_in:NnF \g_cflist_loaded { #1 } {
		\seq_if_in:NnF \g_cflist_pending { #1 } {
			\seq_gput_right:Nn \g_cflist_pending { #1 }
		}
	}
}

\NewDocumentCommand{\cfconsiderloaded}{ m }{
	\seq_gput_right:Nn \g_cflist_loaded {#1}
}

\NewDocumentCommand{\cfremove}{ m }
{
	\seq_gremove_all:Nn \g_cflist_pending { #1 }
}

\NewDocumentCommand{\cfload}{ o }
{
	\seq_if_empty:NTF \g_cflist_pending {\unskip} {
		(cf.\ \cref{\seq_use:Nn \g_cflist_pending {,}})\IfValueTF{#1}{#1~}{\unskip}
		\seq_gconcat:NNN \g_cflist_loaded \g_cflist_loaded \g_cflist_pending
		\seq_gclear:N \g_cflist_pending
	}
}

\NewDocumentCommand{\cfclear} {} {
	\seq_gclear:N \g_cflist_loaded
	\seq_gclear:N \g_cflist_pending
}

\NewDocumentCommand{\cfout}{ o }
{
	\seq_if_empty:NTF \g_cflist_pending {\unskip} {
		(cf.\ \cref{\seq_use:Nn \g_cflist_pending {,}})\IfValueTF{#1}{#1~}{\unskip}
		\seq_gclear:N \g_cflist_pending
	}
}

\NewDocumentCommand{\ifnocf} { m } {
	\seq_if_empty:NT \g_cflist_pending { #1 }
}

\ExplSyntaxOff

\NewDocumentEnvironment{cproof}{m}
{\begin{proof}[Proof of \cref{#1}]}%
	{\noindent The proof of \cref{#1} is thus complete.
\end{proof}}

\NewDocumentEnvironment{cproof2}{m}
{\begin{proof}[Proof of \cref{#1}]}%
	{\noindent This completes the proof of \cref{#1}.
\end{proof}}

\title{Existence, uniqueness, and convergence rates for gradient flows in the training of artificial neural networks with ReLU activation }

\author{Simon Eberle$^1$, Arnulf Jentzen$^{2, 3}$,
		Adrian Riekert$^4$, and Georg S.~Weiss$^5$
		\bigskip
		\\
		\small{$^1$ Faculty of Mathematics,  AG Analysis of Partial Differential Equations,}
		\vspace{-0.1cm}\\
		\small{University of Duisburg-Essen, Germany, e-mail: \texttt{simon.eberle}\textcircled{\texttt{a}}\texttt{uni-due.de}}
		\smallskip
		\\
		\small{$^2$ Applied Mathematics: Institute for Analysis and Numerics,}
		\vspace{-0.1cm}\\
		\small{University of Münster, Germany, e-mail: \texttt{ajentzen}\textcircled{\texttt{a}}\texttt{uni-muenster.de}}
		\smallskip
		\\
		\small{$^3$ School of Data Science and Shenzhen Research Institute of Big Data,}
		\vspace{-0.1cm}\\
		\small{The Chinese University of Hong Kong, Shenzhen, China, e-mail: \texttt{ajentzen}\textcircled{\texttt{a}}\texttt{cuhk.edu.cn}}
		\smallskip
		\\
		\small{$^4$ Applied Mathematics: Institute for Analysis and Numerics,}
		\vspace{-0.1cm}\\
		\small{University of Münster, Germany, e-mail: \texttt{ariekert}\textcircled{\texttt{a}}\texttt{uni-muenster.de}}
		\smallskip
		\\
		\small{$^5$ Faculty of Mathematics,  AG Analysis of Partial Differential Equations,}
		\vspace{-0.1cm}\\
		\small{University of Duisburg-Essen, Germany, e-mail: \texttt{georg.weiss}\textcircled{\texttt{a}}\texttt{uni-due.de}}}

\date{\today}

\begin{document}

\maketitle

\begin{abstract}
\noindent   
The training of artificial neural networks (ANNs) with rectified linear unit (ReLU) activation via gradient descent (GD) type optimization schemes is nowadays 
a common industrially relevant procedure which appears, for example, in the context of natural language processing, image processing, 
fraud detection, and game intelligence.
 Although there exist a large number of numerical simulations in which GD type optimization schemes are effectively used to train ANNs with ReLU activation, till this day in the scientific literature there is in general no mathematical convergence analysis which explains the success of GD type optimization schemes 
 in the training of such ANNs. GD type optimization schemes can be regarded as temporal discretization methods for the gradient flow (GF) differential equations associated to the considered optimization problem and, 
 in view of this, it seems to be a natural direction of research 
 to \emph{first aim to develop a mathematical convergence theory for time-continuous GF differential equations} 
 and, thereafter, to aim to extend such a time-continuous convergence theory 
 to implementable time-discrete GD type optimization methods. 
 In this article we establish two basic results for GF differential equations 
 in the training of fully-connected feedforward ANNs with one hidden layer and ReLU activation. 
 In the first main result of this article we establish 
 in the training of such ANNs under the assumption that the probability 
 distribution of the input data of the considered supervised learning problem 
 is absolutely continuous with a bounded density function 
 that every GF differential equation admits for every initial value 
 a solution which is also unique among a suitable class of solutions. 
 In the second main result of this article we prove in the training of such ANNs 
 under the assumption that the target function and the density function 
 of the probability distribution of the input data are piecewise polynomial   
 that every non-divergent GF trajectory converges with an appropriate rate of convergence 
 to a critical point and that the risk of the non-divergent GF trajectory converges with rate 1 to the risk of the critical point.
\end{abstract}

\pagebreak 
\tableofcontents

\section{Introduction}

The training of artificial neural networks (ANNs) with rectified linear unit (ReLU) activation via gradient descent (GD) type optimization schemes is nowadays a common industrially relevant procedure which appears, for instance, in the context of natural language processing, face recognition, 
fraud detection, and game intelligence.
Although there exist a large number of numerical simulations in which GD type optimization schemes are effectively used to train ANNs with ReLU activation, till this day in the scientific literature there is in general no mathematical convergence analysis which explains the success of GD type optimization schemes in the training of such ANNs.

GD type optimization schemes can be regarded as temporal discretization methods for the gradient flow (GF) differential equations associated to the considered optimization problem and, in view of this, it seems to be a natural direction of research to \emph{first aim to develop a mathematical convergence theory for time-continuous GF differential equations} and, thereafter, to aim to extend such a time-continuous convergence theory to implementable time-discrete GD type optimization methods.

Although there is in general no theoretical analysis which explains the success of GD type optimization schemes in the training of ANNs in the literature, there are several auspicious analysis approaches as well as several promising partial error analyses regarding the training of ANNs via GD type optimization schemes and GFs, respectively, in the literature. 
For convex objective functions,
the convergence of GF and GD processes to the global minimum in different settings has been proved, e.g., in~\cite{BachMoulines2013, JentzenKuckuckNeufeldVonWurstemberger2021, BachMoulines2011, Nesterov2004, Rakhlin2012}.
For general non-convex objective functions, even under smoothness assumptions GF and GD processes can show wild oscillations and admit infinitely many limit points, cf., e.g.,~\cite{AbsilMahonyAndrews2005}.
A standard condition which excludes this undesirable behavior is the \L ojasiewicz inequality and we point to~\cite{AbsilMahonyAndrews2005,AttouchBolte2009,AttouchBolteSvaiter2013,BolteDaniilidis2006,DereichKassing2021,Karimi2020linear,KurdykaMostowski2000,LeePanageasRecht2019,LeeJordanRecht2016,Lojasiewicz1984,Ochs2019} for convergence results for GF and GD processes under \L ojasiewicz type assumptions.
It is in fact one of the main contributions of this work to demonstrate that the objective functions occurring in the training of ANNs with ReLU activation satisfy an appropriate \L ojasiewicz inequality,
provided that both the target function and the density of the probability distribution of the input data are piecewise polynomial.
For further abstract convergence results for GF and GD processes
in the non-convex setting we refer, e.g., to~\cite{BertsekasTsitsiklis2000, FehrmanGessJentzen2020,LeiHuLiTang2020,Patel2021stopping,Santambrogio2017} and the references mentioned therein.

In the overparametrized regime, where the number of training parameters is much larger than the number of training data points, GF and GD processes can be shown to converge to global minima in the training of ANNs with high probability, cf., e.g.,~\cite{AroraDuHuLiWang2019, BachChizatOyallon2019,DuZhaiPoczosSingh2019,EMaWu2020,JacotGabrielHongler2018,JentzenKroeger2021,ZhangMartensGrosse2019}.
As the number of neurons increases to infinity, the corresponding GF processes converge (with appropriate rescaling) to a measure-valued process which is known in the scientific literature as Wasserstein gradient flow.
For results on the convergence behavior of Wasserstein gradient flows in the training of ANNs we point, e.g., to~\cite{ZhengdaoRotskoff2020}, \cite{Chizat2021}, \cite{ChizatBach2018}, \cite[Section 5.1]{EMaWojtowytschWu2020}, and the references mentioned therein.

A different approach is to consider only very special target functions
and we refer, in particular, to \cite{CheriditoJentzenRiekert2021,JentzenRiekert2021} for a convergence analysis for GF and GD processes in the case of constant target functions and to \cite{JentzenRiekert2021rates} for a convergence analysis for GF and GD processes in the training of ANNs with piecewise linear target functions.
In the case of linear target functions, a complete characterization of the non-global local minima and the saddle points of the risk function has been obtained in \cite{Cheridito2021landscape}.

In this article we establish two basic results for GF differential equations in the training of fully-connected feedforward ANNs with one hidden layer and ReLU activation. 
Specifically, in the first main result of this article, 
see \cref{theo:intro:existence} below, 
we establish in the training of such ANNs 
under the assumption that the probability distribution of the input data 
of the considered supervised learning problem is absolutely continuous 
with a bounded density function 
that every GF differential equation possesses 
for every initial value a solution which 
is also unique among a suitable class of solutions 
(see \cref{theo:intro:exist:eq:gf} in \cref{theo:intro:existence} for details). 
In the second main result of this article, see \cref{theo:intro:convergence} below, 
we prove in the training of such ANNs under the assumption that the 
target function and the density function are piecewise polynomial 
(see \cref{theo:intro:conv:eq0} below for details) 
that every non-divergent GF trajectory converges 
with an appropriate speed of convergence (see \cref{theo:intro:conv:eq3} below) to a critical point.

In \cref{theo:intro:existence,theo:intro:convergence} 
we consider ANNs with $ d \in \N = \{ 1, 2, 3, \dots \} $ neurons on the input layer ($d$-dimensional input), $H \in \N$ neurons on the hidden layer ($ H $-dimensional hidden layer), and $1$ neuron on the output layer ($ 1 $-dimensional output). There are thus $H d$ scalar real weight parameters and $H$ scalar real bias parameters to describe the affine linear transformation between $d$-dimensional input layer and the $H$-dimensional hidden layer and there are thus $H$ scalar real weight parameters and 1 scalar real bias parameter to describe the affine linear transformation between the $ H $-dimensional hidden layer and the $ 1 $-dimensional output layer. Altogether there are thus 
$\fd = H d + H + H + 1 = H d + 2 H + 1$
real numbers to describe the ANNs in \cref{theo:intro:existence,theo:intro:convergence}.

The real numbers $\scra \in \R$, $\scrb \in (\scra , \infty )$ in \cref{theo:intro:existence,theo:intro:convergence} are used to specify the set $ [ \scra , \scrb ]^d$ in which the input data of the considered supervised learning problem takes values in and the 
function $f \colon [\scra , \scrb ]^d \to \R$ in \cref{theo:intro:existence} specifies the target function of the considered supervised learning problem. 

In \cref{theo:intro:existence} we assume that the target function is an element of the set $C( [\scra , \scrb ]^d, \R )$ of continuous functions from $[\scra , \scrb]^d$ to $\R$ but beside this continuity hypothesis we do not impose further regularity assumptions on the target function. 

The function $\dens \colon [ \scra , \scrb ]^d \to [0 , \infty )$ in \cref{theo:intro:existence,theo:intro:convergence} is an unnormalized density function of the probability distribution of the input data of the considered supervised learning problem and in \cref{theo:intro:existence} we impose that this unnormalized density function is bounded and measurable.

In \cref{theo:intro:existence,theo:intro:convergence} we consider ANNs with the ReLU activation function $\R \ni x \mapsto \max \cu{ x, 0 } \in \R$. The ReLU activation function fails to be differentiable and this lack of regularity also transfers to the risk function of the considered supervised learning problem; cf.~\cref{theo:intro:general:eq2} below. We thus need to employ appropriately generalized gradients of the risk function to specify the dynamics of the gradient flows. 
As in \cite[Setting~2.1 and Proposition~2.3]{JentzenRiekert2021} 
(cf.\ also \cite{CheriditoJentzenRiekert2021,JentzenRiekertFlow}), 
we accomplish this, first, by approximating 
the ReLU activation function through continuously differentiable functions which converge pointwise to the ReLU activation function and whose derivatives converge pointwise to the left derivative of the ReLU activation function and, thereafter, by specifying the generalized gradient function as the limit of the gradients of the approximated risk functions; see \cref{theo:intro:general:eq1,theo:intro:general:eq2} in \cref{theo:intro:existence} and \cref{theo:intro:conv:eq1,theo:intro:conv:eq2} in \cref{theo:intro:convergence} for details. 

We now present the precise statement of \cref{theo:intro:existence} and, 
thereafter, provide further comments regarding \cref{theo:intro:convergence}.

\begin{samepage}
\begin{theorem} \label{theo:intro:existence}
Let $d, \width, \fd \in \N$, $ \scra \in \R$, $\scrb \in ( \scra, \infty)$, $f \in C ( [\scra , \scrb ]^d , \R)$ satisfy $\fd = d\width + 2 \width + 1$,
let $\dens  \colon  [\scra , \scrb]^d  \to [0, \infty)$ be bounded and measurable,
let $\Rect_r \in C ( \R , \R )$, $r \in \N \cup \cu{ \infty } $, satisfy for all $x \in \R$ that $( \bigcup_{r \in \N} \cu{ \Rect_r } ) \subseteq C^1( \R , \R)$, $\Rect_\infty ( x ) = \max \cu{ x , 0 }$,
 $\sup_{r \in \N} \sup_{y \in [- \abs{x}, \abs{x} ] } \abs{ ( \Rect_r)'(y)} < \infty$, and
\begin{equation} \label{theo:intro:general:eq1}
    \limsup\nolimits_{r \to \infty}  \rbr*{ \abs { \Rect_r ( x ) - \Rect _\infty ( x ) } + \abs { (\Rect_r)' ( x ) - \indicator{(0, \infty)} ( x ) } } = 0,
\end{equation}
for every $\theta = ( \theta_1, \ldots, \theta_\fd) \in \R^\fd$ let $\bfD ^\theta \subseteq \N$ satisfy 
\begin{equation}
    \bfD^\theta = \big\{ 
      i \in \cu{  1, 2, \ldots, \width } \colon \abs{\theta_{\width d + i } } + \smallsum_{j=1}^d \abs{\theta_{(i - 1 ) d + j } } = 0 
    \big\},
\end{equation}
let $\cL_r \colon \R^\fd \to \R$, $r \in \N \cup \cu{ \infty }$,
satisfy for all $r \in \N \cup \cu{ \infty }$, $\theta = (\theta_1, \ldots, \theta_\fd) \in \R^{\fd}$ that
\begin{multline} \label{theo:intro:general:eq2}
   \cL_r ( \theta ) 
   = \int_{[\scra , \scrb]^d} \bigl( f ( x_1, \ldots, x_d )  \\ - \theta_{\fd} - \smallsum_{i=1}^\width \theta_{\width ( d + 1 ) + i } \br[\big]{ \Rect_r ( \theta_{\width d + i}  + \smallsum_{j=1}^d\theta_{(i-1)d + j } x_j ) } \bigr) ^2  \dens ( x ) \,\d (x_1, \ldots, x_d ) ,
    \end{multline}
let $ \theta \in \R^\fd $, and 
let $\cG  \colon \R^\fd \to \R^\fd$ satisfy for all
$
  \vartheta
  \in \cu{ v \in \R^\fd \colon ( ( \nabla \cL_r ) ( v ) ) _{r \in \N} \text{ is convergent} }
$
that 
$
  \cG ( \vartheta ) = \allowbreak \lim_{r \to \infty} \allowbreak (\nabla \cL_r) ( \vartheta )
$.
Then
\begin{enumerate} [label = (\roman*)]
    \item it holds that $\cG$ is locally bounded and measurable and
    \item there exists a unique $\Theta \in C([0, \infty ) , \R^\fd)$ which satisfies for all $t \in [0, \infty)$, $s \in [t, \infty)$ that $\bfD^{\Theta_t} \subseteq \bfD^{\Theta_s }$ and
\begin{equation} \label{theo:intro:exist:eq:gf}
    \Theta_t = \theta - \int_0^t \cG ( \Theta_u ) \, \d u .
\end{equation}
\end{enumerate}
\end{theorem}
\end{samepage}

\cref{theo:intro:existence} is a direct consequence of \cref{theo:gf:exist:unique} below.
 In \cref{theo:intro:convergence} we also assume that the target function $f \colon [ \scra , \scrb]^d \to \R$ is continuous but additionally assume that, 
 roughly speaking,
  both the target function $f \colon [ \scra , \scrb]^d \to \R$ and the unnormalized density function $\dens \colon [ \scra , \scrb]^d \to [0 , \infty )$ coincide with polynomial functions on suitable subsets of their domain of definition $[ \scra , \scrb]^d$.
  In \cref{theo:intro:convergence} 
  the $ ( n \times d ) $-matrices $ \alpha^k_i \in \R^{ n \times d } $, $ i \in \cu{ 1, 2, \ldots, n } $, $ k \in \cu{0 , 1} $,
   and the $ n $-dimensional vectors $ \beta^k_i \in \R^n $,
     $ i \in \cu{ 1, 2, \ldots, n } $, $ k \in \cu{0 , 1} $,
      are used to describe these subsets and the functions $P^k_i \colon \R^d \to \R$,  $ i \in \cu{ 1, 2, \ldots, n } $, $ k \in \cu{0 , 1} $,
       constitute the polynomials with which the target function 
       and the unnormalized density function should partially coincide. 
       More formally, in \cref{theo:intro:conv:eq0} in \cref{theo:intro:convergence} 
       we assume that for every $x \in [ \scra , \scrb ] ^d$ we have that
\begin{equation}
\textstyle
  \dens (x) = \sum_{i \in \cu{1, 2, \ldots, n }, \,
  \alpha_i^0 x + \beta_i^0 \in [0,\infty)^n } P_i^0 ( x )
\qandq 
  f(x) = 
  \sum_{i \in \cu{1, 2, \ldots, n }, \, 
    \alpha_i^1 x + \beta_i^1 \in [0,\infty)^n 
  } P_i^1 ( x ) .
\end{equation}
In \cref{theo:intro:conv:eq3} in \cref{theo:intro:convergence} we prove that there exists 
a strictly positive real number $\beta \in ( 0 , \infty )$ such that 
for every GF trajectory $\Theta \colon [0,\infty) \to \R^{ \fd }$ which does not diverge to infinity in the sense\footnote{Note that the functions $\norm{ \cdot } \colon ( \cup_{n \in \N} \R ^n ) \to \R$ and $\spro{  \cdot , \cdot } \colon (\cup_{n \in \N} ( \R^n \times \R^n ) ) \to \R$ satisfy for all $n \in \N$, $x = ( x_1, \ldots, x_n )$, $y = ( y_1, \ldots, y_n ) \in \R^n $ that $\norm{ x } = [ \sum_{i=1}^n \abs*{ x_i } ^2 ] ^{1/2}$ and $\spro{ x , y } = \sum_{i=1}^\fd x_i y_i$.} that $\liminf_{t \to \infty} \norm{\Theta_t } < \infty$ we have that $\Theta_t \in \R^{ \fd }$, $t \in [0, \infty) $, converges with order $ \beta $ to a critical point $\vartheta \in \cG^{ - 1 }( \cu{ 0 } ) = \cu{ \theta \in \R^{ \fd } \colon \cG ( \theta ) = 0 }$ and we have that the risk $ \cL ( \Theta_t ) \in \R $, $ t \in [0,\infty) $, converges with order 1 
to the risk $\cL ( \vartheta )$ of the critical point $\vartheta$.
We now present the precise statement of \cref{theo:intro:convergence}.

\begin{samepage}
\begin{theorem} \label{theo:intro:convergence}
Let $d, \width, \fd,  n \in \N$, $ \scra \in \R$, $\scrb \in ( \scra, \infty)$, $f \in C([\scra , \scrb ] ^d , \R)$ satisfy $\fd = d\width + 2 \width + 1$,
for every $i \in \cu{1,2, \ldots, n}$, $k \in \cu{0,1}$ let $\alpha_{i}^k \in \R^{n \times d}$,
let $\beta_{i }^k \in \R^n$, 
and
let $P_i^k \colon \R^d \to \R$ be a polynomial,
let $\dens \colon [\scra , \scrb ] ^d \to [0, \infty)$
satisfy for all $k \in \cu{0,1}$,
$x  \in [\scra , \scrb] ^d$ that
\begin{equation} 
\label{theo:intro:conv:eq0}
    k f ( x ) + ( 1 - k ) \dens ( x ) = \smallsum_{i=1}^n \br*{ P_i^k ( x ) \indicator{[0, \infty )^n} \rbr{ \alpha^k _i x + \beta^k_i } } ,
\end{equation}
let $\Rect_r \in C ( \R , \R )$, $r \in \N \cup \cu{ \infty } $, satisfy for all $x \in \R$ that $( \bigcup_{r \in \N} \cu{ \Rect_r } ) \subseteq C^1( \R , \R)$, $\Rect_\infty ( x ) = \max \cu{ x , 0 }$,
 $\sup_{r \in \N} \sup_{y \in [- \abs{x}, \abs{x} ] } \abs{ ( \Rect_r)'(y)} < \infty$, and
\begin{equation} \label{theo:intro:conv:eq1}
    \limsup\nolimits_{r \to \infty}  \rbr*{ \abs { \Rect_r ( x ) - \Rect _\infty ( x ) } + \abs { (\Rect_r)' ( x ) - \indicator{(0, \infty)} ( x ) } } = 0,
\end{equation}
let $\cL_r \colon \R^\fd \to \R$, $r \in \N \cup \cu{ \infty }$,
satisfy for all $r \in \N \cup \cu{ \infty }$, $\theta = (\theta_1, \ldots, \theta_\fd) \in \R^{\fd}$ that
\begin{multline} \label{theo:intro:conv:eq2}
   \cL_r ( \theta ) 
   = \int_{[\scra , \scrb]^d} \bigl( f ( x_1, \ldots, x_d )  \\ - \theta_{\fd} - \smallsum_{i=1}^\width \theta_{\width ( d + 1 ) + i } \br[\big]{ \Rect_r ( \theta_{\width d + i}  + \smallsum_{j=1}^d\theta_{(i-1)d + j } x_j ) } \bigr)^2  \dens ( x ) \,\d (x_1, \ldots, x_d ) ,
    \end{multline}
let $\cG  \colon \R^\fd \to \R^\fd$ satisfy for all
$\theta \in \cu{ \vartheta \in \R^\fd \colon ( ( \nabla \cL_r ) ( \vartheta ) ) _{r \in \N} 
\text{ is convergent} }$
that $\cG ( \theta ) = \lim_{r \to \infty} \allowbreak (\nabla \cL_r) ( \theta )$,
and let $\Theta \in C ( [ 0 , \infty ) , \R^\fd )$ satisfy $\liminf_{t \to  \infty } \norm{\Theta_t } < \infty$ and $\forall \, t \in [0, \infty ) \colon \Theta_t = \Theta_0 - \int_0^t \cG ( \Theta_s ) \, \d s$.
Then there exist $\vartheta \in \cG^{ - 1 } ( \cu{  0 } )$,
$ \fC,  \beta \in (0, \infty)$ which satisfy for all $t \in [ 0 , \infty )$ that
\begin{equation} \label{theo:intro:conv:eq3}
    \norm{\Theta_t - \vartheta} \leq  \fC ( 1 + t ) ^{- \beta } 
\qqandqq
  \abs{ \cL_\infty ( \Theta_t ) - \cL_\infty ( \vartheta ) } \leq \fC ( 1 + t ) ^{-1} .
\end{equation}
\end{theorem}
\end{samepage}

\cref{theo:intro:convergence} above is an immediate consequence of \cref{theo:gf:conv:simple} in \cref{subsection:gf:global:conv} below. \cref{theo:intro:convergence} is related to Theorem 1.1 in our previous article \cite{JentzenRiekertFlow}. 
In particular, \cite[Theorem 1.1]{JentzenRiekertFlow} uses weaker assumptions than \cref{theo:intro:convergence} above but \cref{theo:intro:convergence} above establishes a stronger statement when compared to \cite[Theorem 1.1]{JentzenRiekertFlow}.
 Specifically, on the one hand in \cite[Theorem 1.1]{JentzenRiekertFlow} the target function is only assumed to be a continuous function and the unnormalized density is only assumed to be measurable and integrable while in \cref{theo:intro:convergence} it is additionally assumed that both the target function and the unnormalized density are piecewise polynomial in the sense of \cref{theo:intro:conv:eq0} above.  
On the other hand \cite[Theorem 1.1]{JentzenRiekertFlow} only asserts that the risk of every bounded GF trajectory converges to the risk of critical point while \cref{theo:intro:convergence} assures that every non-divergent GF trajectory converges with a polynomial rate of convergence to a critical point and also assures that the risk of the non-divergent GF trajectory converges with rate 1 to the risk of the critical point.

The remainder of this article is organized in the following way. 
In \cref{section:risk:diff} we establish several regularity properties for the risk function of the considered supervised learning problem and its generalized gradient function. 
In \cref{section:gf:existence} we employ the findings from \cref{section:risk:diff} to establish existence and uniqueness properties for solutions of GF differential equations. 
In particular, in \cref{section:gf:existence} we present the proof of \cref{theo:intro:existence} above. 
In \cref{section:semialgebraic} we establish under the assumption that both the target function $f \colon [\scra , \scrb ]^d \to \R$ and the unnormalized density function $\dens \colon [\scra , \scrb ]^d \to [ 0 , \infty )$ are piecewise polynomial 
that the risk function is semialgebraic in the sense of \cref{def:semialgebraic:function} in \cref{section:semialgebraic} (see \cref{cor:loss:semialgebraic} in \cref{section:semialgebraic} for details). 
In \cref{section:gf:loja} we engage the results from \cref{section:risk:diff,section:semialgebraic} to establish several convergence rate results for solutions of GF differential equations 
and, thereby, we also prove \cref{theo:intro:convergence} above.

\section{Properties of the risk function and its generalized gradient function}
\label{section:risk:diff}

In this section we establish several regularity properties 
for the risk function $\cL \colon \R^\fd \to \R$ and 
its generalized gradient function $\cG \colon \R^\fd \to \R^\fd$.
In particular, in \cref{prop:loss:gradient:subdiff} in \cref{subsection:subdiff} below 
we prove for every parameter vector $ \theta \in \R^{ \fd } $ in the ANN parameter space $ \R^{ \fd } = \R^{ d H  + 2 H + 1 } $ 
that the generalized gradient $ \cG ( \theta ) $ is a limiting subdifferential 
of the risk function $\cL \colon \R^{ \mathfrak{d} } \to \R$ at $ \theta $.
In \cref{def:subdifferential} in \cref{subsection:subdiff} 
we recall the notion of subdifferentials (which are sometimes also 
referred to as Fr\'{e}chet subdifferentials in the scientific literature) 
and in \cref{def:limit:subdiff} in \cref{subsection:subdiff} we recall the notion of limiting subdifferentials.
In the scientific literature
\cref{def:subdifferential,def:limit:subdiff} can in a slightly different presentational form, e.g., be found in Rockafellar \& Wets~\cite[Definition 8.3]{RockafellarWets1998} and Bolte et al.~\cite[Definition 2.10]{BolteDaniilidis2006}, respectively.

Our proof of \cref{prop:loss:gradient:subdiff} uses the continuously differentiability result for the risk function in \cref{prop:loss:continuously:diff} in \cref{subsection:risk:diff} and the local Lipschitz continuity result for the generalized gradient function in \cref{cor:gradient:comp:lip} in \cref{subsection:gradient:lip}.
\cref{cor:gradient:comp:lip} will also be employed in \cref{section:gf:existence} below to establish existence and uniqueness results for solutions of GF differential equations. 
 \cref{prop:loss:continuously:diff} follows directly from  \cite[Proposition 2.11, Lemma 2.12, and Lemma 2.13]{JentzenRiekertFlow}.
  Our proof of \cref{cor:gradient:comp:lip}, in turn, employs the known representation result for the generalized gradient function in \cref{prop:loss:approximate:gradient} in \cref{subsection:risk:diff} below and the local Lipschitz continuity result for certain parameter integals in \cref{cor:derivative:integral:lip} in \cref{subsection:gradient:lip}. 
  Statements related to \cref{prop:loss:approximate:gradient} can, e.g., be found in \cite[Proposition 2.2]{JentzenRiekertFlow}, \cite[Proposition 2.3]{CheriditoJentzenRiekert2021},
 and \cite[Proposition 2.3]{JentzenRiekert2021}.

Our proof of \cref{cor:derivative:integral:lip} uses the elementary abstract local Lipschitz continuity result for certain parameter integrals in \cref{lem:integral:interval:lipschitz} in \cref{subsection:gradient:lip} and the local Lipschitz continuity result for active neuron regions in \cref{lem:active:intervals} in \cref{subsection:neuron:regions} below.
\cref{lem:active:intervals} is a generalization of \cite[Lemma 2.8]{JentzenRiekert2021rates},
  \cref{lem:integral:interval:lipschitz} is a slight generalization of \cite[Lemma 2.7]{JentzenRiekert2021rates},
   and \cref{cor:derivative:integral:lip} is a generalization of \cite[Lemma 2.13]{JentzenRiekertFlow} and \cite[Corollaries 2.10 and 2.11]{JentzenRiekert2021rates}. Only for completeness we include in this section a detailed proof for \cref{lem:integral:interval:lipschitz}.
   In \cref{setting:snn} in \cref{subsection:setting:snn} below we present the mathematical setup to describe ANNs with ReLU activation, the risk function $\cL \colon \R^\fd \to \R$, and its generalized gradient function $\cG \colon \R^\fd \to \R^\fd$. 
   Moreover, in \cref{setting:eq:degen} in \cref{setting:snn} we define for a given parameter vector $\theta \in \R^\fd$ the set of hidden neurons which have all input parameters equal to zero. Such neurons are sometimes called degenerate (cf.~\cite{Cheridito2021landscape}) and can cause problems with the differentiability of the risk function,
   which is why we exclude degenerate neurons in \cref{prop:loss:continuously:diff,cor:gradient:comp:lip} below.

\subsection{Mathematical description of artificial neural networks (ANNs)}
\label{subsection:setting:snn}

\begin{setting} \label{setting:snn}
Let $d, \width, \fd \in \N$, $ \scra \in \R$, $\scrb \in ( \scra, \infty)$, $f \in C ( [\scra , \scrb ]^d , \R)$ satisfy $\fd =  d \width + 2 \width + 1$,
let $\fw  = (( \w{\theta} _ {i,j}  )_{(i , j ) \in \cu{ 1, \ldots, \width } \times \cu{  1, \ldots, d } })_{ \theta \in \R^{\fd}} \colon \R^{\fd} \to \R^{\width \times d}$,
$\fb =  (( \b{\theta} _ 1 , \ldots, \b{\theta} _ \width ))_{ \theta \in \R^{\fd}} \colon \R^{\fd} \to \R^{\width}$,
$\scrb = (( \v{\theta} _ 1 , \ldots, \v{\theta} _ \width ))_{ \theta \in \R^{\fd}} \colon \R^{\fd} \to \R^{\width}$, and
$\fc = (\c{\theta})_{\theta \in \R^{\fd }} \colon \R^{\fd} \to \R$
 satisfy for all $\theta  = ( \theta_1 ,  \ldots, \theta_{\fd}) \in \R^{\fd}$, $i \in \cu{ 1, 2, \ldots, \width }$, $j \in \cu{ 1, 2, \ldots, d }$ that 
 \begin{equation}
     \w{\theta}_{i , j } = \theta_{ (i - 1 ) d + j}, \qquad \b{\theta}_i = \theta_{\width d + i}, \qquad 
\v{\theta}_i = \theta_{  \width (d + 1 ) + i}, \qqandqq \c{\theta} = \theta_{\fd},
 \end{equation}
let $\Rect_r \in C^1 ( \R , \R )$, $r \in \N $, satisfy for all $x \in \R$ that 
\begin{equation}
    \limsup\nolimits_{r \to \infty}  \rbr*{ \abs { \Rect_r ( x ) - \max \cu{  x , 0} } + \abs { (\Rect_r)' ( x ) - \indicator{(0, \infty)} ( x ) } } = 0
\end{equation}
and $\sup_{r \in \N} \sup_{y \in [- \abs{x}, \abs{x} ] }  \abs{(\Rect_r)'(y)} < \infty$,
let $\lambda \colon \cB ( \R^d ) \to [0, \infty ]$ be the Lebesgue--Borel measure on $\R^d$,
let $\dens \colon [\scra , \scrb ] ^d \to [0, \infty)$ be bounded and measurable,
let $\scrN = (\realization{\theta})_{\theta \in \R^{\fd } } \colon \R^{\fd } \to C(\R^d , \R)$ and $\cL \colon \R^{\fd  } \to \R$
satisfy for all $\theta \in \R^{\fd}$, $x =(x_1 , \ldots, x_d) \in \R^d$ that 
\begin{equation} 
\realization{\theta} (x) = \c{\theta} + \smallsum_{i=1}^\width \v{\theta}_i \max \cu[\big]{ \b{\theta}_i + \smallsum_{j=1}^d \w{\theta}_{i , j} x_j , 0 }
\end{equation} 
and $\cL (\theta) = \int_{[ \scra , \scrb ] ^d} ( f ( y ) - \realization{\theta} (y) )^2 \dens ( y ) \, \lambda ( \d y ) $,
let $\fL_r \colon \R^\fd \to \R$, $r \in \N$,
satisfy for all $r \in \N$, $\theta \in \R^{\fd}$ that
\begin{equation}
    \fL_r ( \theta ) = \int_{ [ \scra , \scrb ] ^d} \rbr*{f(y) - \c{\theta} - \smallsum_{i = 1}^\width \v{\theta}_i \br[\big]{ \Rect_r \rbr[\big]{ \b{\theta}_i + \smallsum_{j=1}^d \w{\theta}_{i , j } y _j } }}^{ \! 2 } \dens ( y ) \, \lambda ( \d y ),
\end{equation}
for every $\varepsilon \in (0, \infty)$, $\theta \in \R^\fd$ let $B_\varepsilon ( \theta ) \subseteq \R^\fd$ satisfy $B_\varepsilon ( \theta ) = \cu{\vartheta \in \R^\fd \colon \norm{\theta - \vartheta } < \varepsilon }$,
for every $\theta \in \R^{\fd }$, $i \in \cu{ 1, 2, \ldots, \width }$
let $I_i^\theta \subseteq \R^d$ satisfy 
\begin{equation} \label{setting:eq:active:int}
    I_i^\theta = \cu[\big]{ x = ( x_1, \ldots, x_d ) \in [\scra , \scrb ]^d \colon \b{\theta}_i +  \smallsum_{j=1}^d  \w{\theta}_{i , j} x_d > 0 },
\end{equation}
for every $\theta \in \R^\fd$ let $\bfD ^\theta \subseteq \N $ satisfy 
\begin{equation} \label{setting:eq:degen}
    \bfD^\theta = \cu[\big]{ i \in \cu{  1, 2, \ldots, \width } \colon \abs{\b{\theta}_i} + \smallsum_{j=1}^d \abs{\w{\theta}_{i , j} } = 0 },
\end{equation}
and let $\cG = ( \cG_1 , \ldots, \cG_\fd ) \colon \R^\fd \to \R^\fd$ satisfy for all $\theta \in  \cu{  \vartheta \in \R^\fd \colon ((\nabla \fL_r)(\vartheta ) )_{r \in \N} \text{ is convergent} }$ that $\cG ( \theta ) = \lim_{r \to \infty} ( \nabla \fL_r) ( \theta ) $.

\end{setting}

\subsection{Differentiability properties of the risk function}
\label{subsection:risk:diff}

\begin{prop} \label{prop:loss:approximate:gradient}
Assume \cref{setting:snn}. Then it holds for all $\theta \in \R^{\fd}$, $i \in \cu{ 1, 2, \ldots, \width }$, $j \in \cu{ 1, 2, \ldots, d }$ that
\begin{equation} \label{eq:loss:gradient}
\begin{split}
        \cG_{ ( i - 1 ) d + j } ( \theta) &= 2 \v{\theta}_i \int_{I_i^\theta} x_j  ( \realization{\theta} (x) - f ( x ) ) \dens ( x ) \,  \lambda ( \d x ) , \\
        \cG_{ \width d + i} ( \theta) &= 2 \v{\theta}_i \int_{I_i^\theta} (\realization{\theta} (x) - f ( x ) ) \dens ( x ) \, \lambda( \d x ) , \\
        \cG_{\width (d + 1 ) + i} ( \theta) &= 2 \int_{[ \scra , \scrb ] ^d} \br[\big]{ \max \cu[\big]{ \b{\theta}_i + \smallsum_{j=1}^d \w{\theta}_{i,j} x_j , 0 } }  ( \realization{\theta}(x) - f ( x ) ) \dens ( x ) \, \lambda ( \d x ), \\
       \text{and} \qquad \cG_{\fd} ( \theta) &= 2 \int_{[ \scra , \scrb ] ^d} (\realization{\theta} (x) - f ( x ) ) \dens ( x ) \, \lambda ( \d x ) .
        \end{split}
\end{equation}
\end{prop}
\begin{cproof}{prop:loss:approximate:gradient}
\Nobs that, e.g., \cite[Proposition 2.2]{JentzenRiekertFlow} establishes \cref{eq:loss:gradient}.
\end{cproof}

\begin{prop} \label{prop:loss:continuously:diff}
Assume \cref{setting:snn} and let $U \subseteq \R^\fd$ satisfy $U = \cu[\big]{\theta \in \R^\fd \colon \bfD^\theta = \emptyset }$.
Then
\begin{enumerate} [label = (\roman*)]
    \item \label{prop:loss:continuously:diff:item1} it holds that $U \subseteq \R^\fd$ is open,
    \item \label{prop:loss:continuously:diff:item2} it holds that $\cL | _U \in C^1 ( U , \R)$, and
    \item \label{prop:loss:continuously:diff:item3} it holds that $\nabla ( \cL |_U ) = \cG |_U$.
\end{enumerate}
\end{prop}
\begin{cproof}{prop:loss:continuously:diff}
\Nobs that \cite[Proposition 2.11, Lemma 2.12, and Lemma 2.13]{JentzenRiekertFlow} establish \cref{prop:loss:continuously:diff:item1,prop:loss:continuously:diff:item2,prop:loss:continuously:diff:item3}.
\end{cproof}

\subsection{Local Lipschitz continuity of active neuron regions}
\label{subsection:neuron:regions}

\cfclear
\begin{lemma} \label{lem:active:intervals} 
	Let $d \in \N$,
	$\scra \in \R$, $\scrb \in (\scra, \infty)$,
	for every $v = ( v_1 , \ldots, v_{d+1} ) \in \R^{d+1}$
	let $I^v \subseteq [\scra ,\scrb ]^d$ satisfy
	$I^v = \cu{  x \in [\scra , \scrb ] ^d \colon v_{d+1} + \smallsum_{i=1}^d v_i x_i > 0 }$,
	for every $n \in \N$ let $\lambda_n \colon \cB ( \R^n) \to [0, \infty]$ be the Lebesgue--Borel measure on $\R^n$,
	let $\dens \colon [\scra , \scrb ] ^d \to [0, \infty )$ be bounded and measurable,
	and let $u \in \R^{d+1} \backslash \cu{0}$.
	Then there exist $\varepsilon , \fC \in (0, \infty)$ such that for all $ v , w \in \R^{d+1}$ with $\max \cu{  \norm{u - v} , \norm{u - w } } \le \varepsilon$ it holds that 
	\begin{equation} \label{lem:active:intervals:eq}
	\textstyle\int_{ I^v \Delta I^w } \dens ( x ) \, \lambda_d ( \d x) \leq \fC \norm{v - w }.
	\end{equation}
\end{lemma}
\begin{cproof} {lem:active:intervals}
	\Nobs that for all $v, w \in \R^{d+1}$ we have that
	\begin{equation} \label{lem:active:intervals:eq:help0}
	\textstyle\int_{ I^v \Delta I^w } \dens ( x ) \, \lambda_d ( \d x) \leq \rbr[\big]{\sup \nolimits_{x \in [\scra , \scrb ]^d} \dens ( x ) } \lambda_d ( I^v \Delta I^w ).
	\end{equation}
	Moreover, \nobs that the fact that for all $y \in \R$ it holds that $y \geq - \abs{y}$ ensures that for all $v = ( v_1, \ldots, v_{d+1} ) \in \R^{d+1}$, $i \in \cu{1, 2, \ldots, d+1 }$ with $\norm{u-v} < \abs{u_i}$ it holds that
	\begin{equation} \label{lem:active:intervals:eq:help1}
	u_i v_i = (u_i)^2 + ( v_i - u_i ) u_i \geq \abs{u_i}^2 - \abs{u_i - v_i} \abs{u_i} \geq \abs{u_i}^2 - \norm{u-v} \abs{u_i} > 0.
	\end{equation}
	Next \nobs that for all $v_1, v_2, w_1, w_2 \in \R$ with $\min \cu{\abs{v_1}, \abs{w_1}} > 0$ it holds that
	\begin{equation}
	\abs*{\tfrac{v_2}{v_1} - \tfrac{w_2}{w_1}} = \tfrac{\abs{v_2 w_1 - w_2 v_1}}{\abs{v_1 w_1 }} = \tfrac{\abs{ v_2 ( w_1 - v_1 ) + v_1 ( v_2 - w_2 ) }}{\abs{v_1 w_1 }} \le \br*{ \tfrac{\abs{v_2} + \abs{v_1}}{\abs{v_1 w_1}}} \br[\big]{ \abs{v_1 - w_1} + \abs{v_2 - w_2 }} .
	\end{equation}
	Combining this and \cref{lem:active:intervals:eq:help1} demonstrates for all $v=(v_1, \ldots, v_{d+1})$, $w=(w_1, \ldots, w_{d+1}) \in \R^{d+1}$, $i \in \cu{1, 2, \ldots, d}$ with $\max \cu{\norm{v-u}, \norm{w-u}} < \abs{u_1}$ that $v_1 w_1 > 0$ and
	\begin{equation}
	\abs*{ \tfrac{v_i}{v_1} - \tfrac{w_i}{w_1}} \le \br*{\tfrac{2 \norm{v}}{ \abs{v_1 w_1 }}} [2 \norm{v - w } ] \le \br*{\tfrac{4 \norm{v - u } + 4 \norm{u}}{ \abs{v_1 w_1 }}}  \norm{v - w } .
	\end{equation}
	Hence, we obtain for all $v=(v_1, \ldots, v_{d+1})$, $w=(w_1, \ldots, w_{d+1}) \in \R^{d+1}$, $i \in \cu{1, 2, \ldots, d}$ with $\max \cu{\norm{v-u}, \norm{w-u}} \le \frac{\abs{u_1}}{2}$ and $\abs{u_1} > 0$ that $v_1 w_1 > 0$ and
	\begin{equation} \label{lem:active:intervals:eq:helpnew}
	\abs*{ \tfrac{v_i}{v_1} - \tfrac{w_i}{w_1}} \le \tfrac{(2 \abs{u_1} + 4 \norm{u} ) \norm{v - w } }{\abs{u_1 + (v_1 - u_1)}\abs{u_1 + (w_1 - u_1 ) }}
	\le \tfrac{6 \norm{u} \norm{v - w }}{ ( \abs{u_1} - \norm{v-u} ) ( \abs{u_1} - \norm{w - u } )} \le \br*{ \tfrac{24 \norm{u}}{ \abs{u_1}^2}} \norm{v - w } . 
	\end{equation}
	In the following we distinguish between the case $\max_{i \in \cu{1, 2, \ldots, d } } \abs{u_i} = 0$,  the case $(\max_{i \in \cu{1, 2, \ldots, d } } \abs{u_i} , \allowbreak d ) \in (0, \infty ) \times [ 2 , \infty)$, and the case $(\max_{i \in \cu{1, 2, \ldots, d } } \abs{u_i} , d ) \in (0, \infty ) \times \cu{1}$.
	We first prove \cref{lem:active:intervals:eq} in the case \begin{equation} \label{lem:active:intervals:eq:case1}
	\max\nolimits_{i \in \cu{1, 2, \ldots, d } } \abs{u_i} = 0.
	\end{equation} 
	\Nobs that \cref{lem:active:intervals:eq:case1} and the assumption that $u \in \R^{d+1} \backslash \cu{ 0 }$ imply that $\abs{ u_{d+1} } > 0$.
	Moreover, \nobs that \cref{lem:active:intervals:eq:case1} shows that for all $v = (v_1, \ldots, v_{d+1} ) \in \R^{d+1}$, $x =(x_1, \ldots, x_d ) \in I^u \Delta I^v$ we have that
	\begin{equation} \label{proof:diff:2:eq1}
	\begin{split}
	&\abs[\big]{\rbr[\big]{ \br[\big]{\smallsum_{i=1}^d v_i x_i} + v_{d+1} } - \rbr[\big]{ \br[\big]{\smallsum_{i=1}^d u_i x_i} + u_{d+1} } } \\
	&=\abs[\big]{\br[\big]{ \smallsum_{i=1}^d v_i x_i} + v_{d+1} }  + \abs[\big]{  \br[\big]{ \smallsum_{i=1}^d u_i x_i} + u_{d+1} } 
	\geq \abs[\big]{ \br[\big]{\smallsum_{i=1}^d u_i x_i} + u_{d+1} } = \abs{u_{d+1} }.
	\end{split}
	\end{equation}
	In addition, \nobs that for all $v = (v_1, \ldots, v_{d+1} ) \in \R^{d+1}$, $x =(x_1, \ldots, x_d ) \in [ \scra , \scrb ] ^d$ it holds that
	\begin{equation} \label{proof:diff:2:eq2}
	\begin{split}
	&   \abs[\big]{ \rbr[\big]{ \br[\big]{\smallsum_{i=1}^d v_i x_i} + v_{d+1} } - \rbr[\big]{ \br[\big]{\smallsum_{i=1}^d u_i x_i} + u_{d+1} } } \leq \br[\big]{\smallsum_{i=1}^d  \abs{v_i - u_i} \abs{x_i}  } + \abs{ v_{d+1} - u_{d+1} } \\
	& \leq \max \cu{ \abs{\scra}, \abs{\scrb} } \br[\big]{\smallsum_{i=1}^d  \abs{v_i - u_i}  } + \abs{ v_{d+1} - u_{d+1} } 
	\leq ( 1 + d \max \cu{ \abs{\scra , \scrb } } ) \norm{ v - u }.
	\end{split}
	\end{equation}
	This and \cref{proof:diff:2:eq1} prove that for all $v \in \R^{d+1} $ with $\norm{ u - v } \le \frac{\abs{u_{d+1}}}{2 + d \max \cu{ \abs{\scra , \scrb } }}$ we have that $I^u \Delta I^v = \emptyset$, i.e., $I^u = I^v$.
	Therefore, we get for all $v , w \in \R^{d+1} $ with $\max \cu{ \norm{ u - v } , \norm{u - w} } \le \frac{\abs{u_{d+1}}}{2 + d \max \cu{ \abs{\scra , \scrb } }}$ that $I^v = I^w = I^u$.
	Hence, we obtain for all $v , w \in \R^{d+1} $ with $\max \cu{ \norm{ u - v } , \norm{u - w} } \le \frac{\abs{u_{d+1}}}{2 + d \max \cu{ \abs{\scra , \scrb } }}$ that $\lambda_d(I^v \Delta I^w ) = 0$.
	This establishes \cref{lem:active:intervals:eq} in the case $\max_{i \in \cu{1, 2, \ldots, d } } \abs{u_i} = 0$.
	In the next step we prove \cref{lem:active:intervals:eq} in the case 
	\begin{equation} \label{lem:active:intervals:eq:case2}
	(\max\nolimits_{i \in \cu{1, 2, \ldots, d } } \abs{u_i} , d ) \in (0, \infty ) \times [ 2 , \infty ).
	\end{equation}
	For this we assume without loss of generality that $\abs{ u_1 } > 0$.
	In the following let $J_x^{v, w} \subseteq \R$, $x \in [\scra , \scrb]^{d-1}$,
	$v, w \in \R^{d+1}$,
	satisfy for all $x =(x_2, \ldots, x_d) \in [\scra , \scrb] ^{d-1}$, $v, w \in \R^{d+1}$ that $J_x^{v , w} = \cu{ y \in [\scra , \scrb ] \colon (y, x_2, \ldots, x_d) \in I^v \backslash I^w }$.
	Next \nobs that Fubini's theorem and the fact that for all $v \in \R^{d+1}$ it holds that $I^v$ is measurable show that for all $v , w \in \R^{d+1}$ we have that
	\begin{equation} \label{eq:est:fubini}
	\begin{split}
	&\lambda_d ( I^v \Delta I^w ) = \int_{[\scra , \scrb ]^d } \indicator{I^v \Delta I^w} ( x ) \, \lambda_d ( \d x )
	=
	\int_{[\scra , \scrb ] ^d } \rbr[\big]{ \indicator{I^v \backslash I^w} ( x ) + \indicator{I^w \backslash I^v } ( x ) } \, \lambda_d ( \d x) \\
	&= \int_{[\scra , \scrb]^{d-1} } \int_{[\scra , \scrb ]} \rbr[\big]{ \indicator{I^v \backslash I^w} ( y, x_2, \ldots, x_d ) + \indicator{I^w \backslash I^v } ( y, x_2, \ldots, x_d ) } \, \lambda_1 ( \d y ) \, \lambda_{d-1} (\d ( x_2, \ldots, x_d ) ) \\
	&= \int_{[\scra , \scrb ] ^{d-1} } \int_{[\scra , \scrb ] } \rbr[\big]{ \indicator{J_x^{v , w}} ( y ) + \indicator{J_x^{w , v}} ( y ) } \, \lambda_1 ( \d y ) \, \lambda_{d-1} ( \d x ) \\
	&=
	\int_{[\scra , \scrb ]^{d-1} } ( \lambda_1 ( J_x^{v , w} ) + \lambda_1 ( J_x^{w , v} ) ) \, \lambda_{d-1} ( \d x).
	\end{split}
	\end{equation}
	Furthermore,
	\nobs that for all $x = (x_2, \ldots, x_d) \in [ \scra , \scrb ] ^{d-1}$, $v = (v_1, \ldots, v_{d+1})$, $w = (w_1, \ldots, w_{d+1} ) \in \R^{d+1}$, $\fs \in \cu{ - 1 , 1 }$ with $\min \cu{\fs v_1, \fs w_1 } > 0$ it holds that
	\begin{equation}
	\begin{split}
	J_x^{v , w} &= \cu*{y \in [\scra , \scrb ] \colon
		(y, x_2, \ldots, x_d) \in I^v \backslash I^w }\\
	&= \cu*{ y \in [ \scra , \scrb ] \colon  v_1 y + \br[\big]{ \smallsum_{i=2}^d v_i x_i } + v_{d+1} > 0 \geq w_1 y + \br[\big]{\smallsum_{i = 2}^d w_i x_i } + w_{d+1} }\\
	& = \cu*{y \in [\scra , \scrb ] \colon - \tfrac{\fs }{v_1} \rbr[\big]{ \br[\big]{  \smallsum_{i=2}^d v_i x_i } + v_{d+1} } < \fs y \leq -\tfrac{\fs }{w_1} \rbr[\big]{ \br[\big]{ \smallsum_{i = 2}^d w_i x_i } + w_{d+1} } } .
	\end{split}
	\end{equation}
	Hence, we obtain for all $x = (x_2, \ldots, x_d) \in [ \scra , \scrb ] ^{d-1}$, $v = (v_1, \ldots, v_{d+1})$, $w = (w_1, \ldots, w_{d+1} ) \in \R^{d+1}$, $\fs \in \cu{ - 1 , 1 }$ with $\min \cu{\fs v_1, \fs w_1 } > 0$ that
	\begin{equation} \label{lem:active:intervals:eq:help2}
	\begin{split}
	\lambda_1 ( J_x^{v , w} ) &\leq \abs*{\tfrac{\fs }{v_1} \rbr[\big]{ \br[\big]{ \smallsum_{i=2}^d v_i x_i } + v_{d+1} } -\tfrac{\fs }{w_1} \rbr[\big]{ \br[\big]{ \smallsum_{i=2}^d w_i x_i } + w_{d+1} } } \\
	&\leq \br*{ \smallsum_{i=2}^d \abs[\big]{\tfrac{v_i}{v_1} - \tfrac{w_i}{w_1}} \abs{x_i} } + \abs*{\tfrac{v_{d+1}}{v_1} - \tfrac{w_{d+1}}{w_1} } \\
	&\leq \max \cu{ \abs{\scra} , \abs{\scrb} } \br*{ \smallsum_{i=2}^d  \abs[\big]{\tfrac{v_i}{v_1} - \tfrac{w_i}{w_1}} } + \abs*{\tfrac{v_{d+1}}{v_1} - \tfrac{w_{d+1}}{w_1} }.
	\end{split}
	\end{equation}
	Furthermore, \nobs that \cref{lem:active:intervals:eq:help1} demonstrates for all $v = (v_1, \ldots, v_{d+1}) \in \R^{d+1}$ with $\norm{u - v} < \abs{u_1}$ that $u_1 v_1 > 0$. This implies that for all $v = (v_1, \ldots, v_{d+1})$, $w = (w_1, \ldots, w_{d+1}) \in \R^{d+1}$ with $\max \cu{\norm{u - v} , \norm{u - w} } < \abs{u_1}$ there exists $\fs \in \cu{-1 , 1 }$ such that  $\min \cu{\fs v_1, \fs w_1 } > 0$.
	Combining this and \cref{lem:active:intervals:eq:helpnew} with \cref{lem:active:intervals:eq:help2} proves that there exists $\fC \in \R$ such that for all $x \in [\scra , \scrb]^{d-1}$,
	$v , w \in \R^{d+1}$ with $\max \cu{ \norm{ u - v }, \norm{u - w } } \le \frac{ \abs{ u_1 } }{2}$ we have that $\lambda_1( J_x^{v , w} ) + \lambda_1 ( J_x^{w , v} ) \leq \fC \norm{ v - w }$. This, \cref{eq:est:fubini}, and \cref{lem:active:intervals:eq:help0} establish \cref{lem:active:intervals:eq} in the case $(\max_{i \in \cu{1, 2, \ldots, d } } \abs{u_i} , d ) \in (0, \infty ) \times [ 2 , \infty )$.
	Finally, we prove \cref{lem:active:intervals:eq} in the case 
	\begin{equation} \label{lem:active:intervals:eq:case3}
	(\max\nolimits_{i \in \cu{1, 2, \ldots, d } } \abs{u_i} , d ) \in (0, \infty ) \times \cu{ 1 }.
	\end{equation}
	\Nobs that \cref{lem:active:intervals:eq:case3} demonstrates that $\abs{u_1} > 0$.
	In addition, \nobs that for all $v = (v_1, v_2)$, $w = (w_1, w_2 ) \in \R^{2}$, $\fs \in \cu{ - 1 , 1 }$ with $\min \cu{\fs v_1, \fs w_1 } > 0$ it holds that
	\begin{equation}
	\begin{split}
	I^v \backslash I^w &= \cu*{ y \in [ \scra , \scrb ] \colon v_1 y + v_2 > 0 \geq w_1 y + w_2} = \cu*{ y \in [\scra , \scrb ] \colon - \tfrac{\fs v_2}{v_1} < \fs y \leq - \tfrac{s w_2}{w_1} } \\
	& \subseteq \cu*{ y \in \R \colon - \tfrac{\fs v_2}{v_1} < \fs y \leq - \tfrac{s w_2}{w_1} }.
	\end{split}
	\end{equation}
	Therefore,
	we get for all $v = (v_1, v_2)$, $w = (w_1, w_2 ) \in \R^{2}$, $\fs \in \cu{ - 1 , 1 }$ with $\min \cu{\fs v_1, \fs w_1 } > 0$ that 
	\begin{equation} \label{lem:active:intervals:eq:help3}
	\lambda_1 ( I^v \backslash I^w  ) \leq \abs*{ \rbr*{ - \tfrac{s v_2}{v_1}} - \rbr*{ - \tfrac{\fs w_2}{w_1} } } = \abs*{\tfrac{v_2}{v_1} - \tfrac{w_2}{w_1} }.
	\end{equation}
	Furthermore, \nobs that \cref{lem:active:intervals:eq:help1} ensures for all $v = (v_1, v_2) \in \R^2$ with $\norm{u - v } < \abs{u_1}$ that $u_1 v_1 > 0$. This proves that for all $v = (v_1, v_2)$, $w = (w_1 , w_2 ) \in \R^2$ with $\max \cu{\norm{u - v } , \norm{u - w} } < \abs{u_1}$ there exists $\fs \in \cu{-1 , 1 }$ such that $\min \cu{\fs v_1, \fs w_1 } > 0$.
	Combining this with \cref{lem:active:intervals:eq:help3} demonstrates for all $v = (v_1, v_2)$, $w = (w_1, w_2) \in \R^2$ with $\max \cu{\norm{u - v } , \norm{u - w } }  < \abs{u_1}$ that $\min \cu{ \abs{v_1}, \abs{w_1}} > 0$ and
	\begin{equation}
	\lambda_1 ( I^v \Delta I^w ) = \lambda_1 ( I^v \backslash I^w ) + \lambda_1 ( I^w \backslash I^v ) \leq 2 \abs*{\tfrac{v_2}{v_1} - \tfrac{w_2}{w_1} }.
	\end{equation}
	This,
	\cref{lem:active:intervals:eq:helpnew},
	and \cref{lem:active:intervals:eq:help0} establish \cref{lem:active:intervals:eq} in the case $(\max_{i \in \cu{1, 2, \ldots, d } } \abs{u_i} , d ) \in (0, \infty ) \times \cu{ 1 }$.
\end{cproof}

\subsection{Local Lipschitz continuity properties for the generalized gradient function}
\label{subsection:gradient:lip}

\cfclear
\begin{lemma} \label{lem:integral:interval:lipschitz}
Let $d , n \in \N$, $\scra \in \R$, $\scrb \in (\scra , \infty)$, $x \in \R^n$, $\fC, \varepsilon \in (0, \infty)$, let $\phi \colon \R^n \times [\scra , \scrb]^d \to \R$ be locally bounded and measurable,
assume for all $r \in (0, \infty )$ that
\begin{equation} \label{lem:integral:interval:lip:eqass}
\sup\nolimits_{y , z \in \R^n , \, \norm{y} + \norm{z} \le r , \, y \not= z } \sup\nolimits_{s \in [\scra , \scrb ] ^d } \tfrac{\abs{\phi ( y , s ) - \phi ( z , s ) }}{ \norm{y-z}} < \infty,
\end{equation} 
let $\mu \colon \cB( [ \scra , \scrb ]^d ) \to [0, \infty)$ be a finite measure,
let $I^y \in \cB ([\scra, \scrb]^d )$, $y \in \R^n$, satisfy for all $y , z \in \cu{  v \in \R^n \colon \norm{x-v} \le \varepsilon }$ that $\mu ( I^y \Delta I^z ) \leq \fC \norm{y - z }$,
and let $\Phi \colon \R^n \to \R$ satisfy for all $y \in \R^n$ that
\begin{equation} \label{lem:integral:interval:lip:defphi}
    \Phi ( y ) = \int_{I^y} \phi (y , s ) \, \mu ( \d s ).
\end{equation}
 Then there exists $\scrC \in \R$ such that for all $y , z \in \cu{  v \in \R^n \colon \norm{x - v} \le \varepsilon }$ 
it holds that $\abs{\Phi (y) - \Phi (z) } \leq \scrC \norm{y - z}$.
\end{lemma}
\begin{cproof}{lem:integral:interval:lipschitz}
\Nobs that \cref{lem:integral:interval:lip:eqass} and the assumption that $\phi $ is locally bounded ensure that there exists $\scrC \in \R$ which satisfies for all $y , z \in \cu{  v \in \R^n \colon \norm{x-v} \le \varepsilon }$, $s \in [\scra , \scrb]^d $ with $y \not= z$ that
\begin{equation} \label{lem:integral:interval:lipschitz:eq1}
  \tfrac{  \abs{\phi (y , s) - \phi (z , s) }} {\norm{y - z}} + \abs{\phi (y,s)} + \abs{\phi (z,s)} \leq \scrC. 
\end{equation}
Next \nobs that \cref{lem:integral:interval:lip:defphi} shows for all $y , z \in \R^n$ that
\begin{equation} \label{lem:integral:interval:lipschitz:eq2}
    \abs{\Phi (y) - \Phi (z)} \leq \int_{I^y \cap I^z } \abs{\phi (y,s) - \phi (z,s) } \, \mu ( \d s ) + \int_{I^y \backslash I^z} \abs{\phi (y , s)} \, \mu ( \d s ) + \int_{I^z \backslash I^y} \abs{\phi (z,s)} \, \mu (  \d s ) .
\end{equation}
Moreover, \nobs that \cref{lem:integral:interval:lipschitz:eq1} assures for all $y , z \in \cu{  v \in \R^n \colon \norm{x-v} \le \varepsilon }$ that
\begin{equation} \label{lem:integral:interval:lipschitz:eq3}
    \int_{I^y \cap I^z } \abs{\phi (y , s) - \phi (z , s) } \, \mu ( \d s ) \leq \scrC  \norm{y - z} \mu ( [ \scra , \scrb ] ^d ).
\end{equation}
In the next step we combine \cref{lem:integral:interval:lipschitz:eq1} with the assumption that for all $y , z \in \cu{  v \in \R^n \colon \norm{x-v} \le \varepsilon }$ it holds that $\mu ( I^y \Delta I^z ) \leq \fC \norm{y - z}$ to obtain that for all $y , z \in \cu{  v \in \R^n \colon \norm{x-v} \le \varepsilon }$ it holds that
\begin{equation} 
    \int_{I^y \backslash I^z} \abs{\phi (y,s)} \, \mu( \d s ) + \int_{I^z \backslash I^y} \abs{\phi (z,s)} \, \mu ( \d s ) \leq  \fC \scrC \norm{y - z}.
\end{equation}
This, \cref{lem:integral:interval:lipschitz:eq2}, and \cref{lem:integral:interval:lipschitz:eq3} demonstrate for all $y , z\in \cu{  v \in \R^n \colon \norm{x - v} \le \varepsilon }$ that
\begin{equation}
    \abs{\Phi (y) - \Phi (z)} \leq \scrC ( \fC + \mu ( [ \scra , \scrb ] ^d ) ) \norm{y - z}.
\end{equation}
\end{cproof}

\begin{cor} \label{cor:derivative:integral:lip}
Assume \cref{setting:snn},
let $\phi \colon \R^\fd \times [\scra , \scrb]^d \to \R$ be locally bounded and measurable, and
assume for all $r \in (0, \infty)$ that
\begin{equation} \label{cor:derivative:integral:lip:eqass}
\sup\nolimits_{\theta , \vartheta \in \R^\fd , \, \norm{\theta } + \norm{\vartheta} \le r , \, \theta \not= \vartheta } \sup\nolimits_{x \in [\scra , \scrb ] ^d } \tfrac{\abs{\phi ( \theta , x ) - \phi ( \vartheta , x ) }}{ \norm{\theta - \vartheta }} < \infty.
\end{equation} 
Then
\begin{enumerate} [label = (\roman*)]
    \item \label{cor:derivative:integral:lip:item1} it holds that 
    \begin{equation}
        \R^\fd \ni \theta \mapsto \int_{[ \scra , \scrb ] ^d} \phi (\theta , x ) \dens ( x )  \, \lambda ( \d x ) \in \R
    \end{equation}
    is locally Lipschitz continuous and
    \item \label{cor:derivative:integral:lip:item2} it holds for all $i \in \cu{ 1, 2, \ldots, \width}$ that 
    \begin{equation}
        \cu[\big]{\vartheta \in \R^\fd \colon i \notin \bfD^\vartheta } \ni \theta \mapsto \int_{I_i^\theta } \phi (\theta , x ) \dens ( x ) \, \lambda ( \d x ) \in \R
    \end{equation}
    is locally Lipschitz continuous.
\end{enumerate}
\end{cor}

\cfclear
\begin{cproof} {cor:derivative:integral:lip}
First \nobs that \cref{lem:integral:interval:lipschitz}
(applied for every $\theta \in \R^\fd$ with $n \with \fd$, $x \with \theta$,
$\mu \with ( \cB ( [ \scra , \scrb ] ^d) \ni A \mapsto \int_A \dens ( x ) \, \lambda ( \, \d x ) \in [0,\infty)) $, 
$ (I^y)_{y \in \R^n} \with ([\scra , \scrb]^d)_{y \in \R^\fd}$ in the notation of \cref{lem:integral:interval:lipschitz})
establishes \cref{cor:derivative:integral:lip:item1}.
In the following let
$i \in \cu{1, 2, \ldots, \width}$,
 $\theta \in \cu{\vartheta \in \R^\fd \colon i \notin \bfD^\vartheta } $.
\Nobs that \cref{lem:active:intervals} shows that there exist $ \varepsilon , \fC \in (0, \infty)$ which satisfy for all $\vartheta_1, \vartheta_2 \in   \R^\fd $ with $\max \cu{  \norm{\theta - \vartheta_1}, \norm{\theta - \vartheta_2} } \le \varepsilon$ that 
\begin{equation}
    \textstyle\int_{ I_i^{\vartheta_1} \Delta I_i^{\vartheta_2} } \dens ( x ) \, \lambda ( \d x ) \leq \fC \norm{\vartheta_1 - \vartheta_2 } . 
\end{equation}
Combining this with \cref{lem:integral:interval:lipschitz}
(applied for every $\theta \in \R^\fd$ with $n \with \fd$, $x \with \theta$,
$ \mu \with ( \cB ( [ \scra , \scrb ] ^d) \ni A \mapsto \int_A \dens ( x ) \, \lambda ( \, \d x ) \in [0,\infty)) 
$, $(I^y)_{y \in \R^n} \with (I_i^y)_{ y \in \R^\fd}$ in the notation of \cref{lem:integral:interval:lipschitz}) 
demonstrates that there exists $\scrC \in \R$ such that for all $\vartheta_1, \vartheta_2 \in \R^\fd $ with $\max \cu{  \norm{\theta - \vartheta_1}, \norm{\theta - \vartheta_2} } \le \varepsilon$ it holds that
\begin{equation}
    \abs*{\int_{I_i^{\vartheta_1}} \phi ( \vartheta_1 , x ) \dens ( x ) \, \lambda ( \d x ) - \int_{I_i^{\vartheta_2}} \phi ( \vartheta_2 , x ) \dens ( x ) \, \lambda (\d x ) } \leq \scrC \norm{\vartheta_1 - \vartheta_2 }.
\end{equation}
This establishes \cref{cor:derivative:integral:lip:item2}.
\end{cproof}

\begin{cor} \label{cor:gradient:comp:lip}
Assume \cref{setting:snn}.
Then
\begin{enumerate} [label = (\roman*)]
    \item \label{cor:gradient:comp:lip:item1} it holds for all $k \in \N \cap (  \width d + \width , \fd]$ that 
    \begin{equation}
        \R^\fd \ni \theta \mapsto \cG_k ( \theta ) \in \R
    \end{equation}
    is locally Lipschitz continuous,
    \item \label{cor:gradient:comp:lip:item2} it holds for all $i \in \cu{ 1, 2, \ldots, \width}$, $j \in \cu{  1, 2, \ldots, d}$ that 
    \begin{equation}
        \cu[\big]{\vartheta \in \R^\fd \colon i \notin \bfD^\vartheta } \ni \theta \mapsto \cG_{(i - 1 ) d + j } ( \theta ) \in \R
    \end{equation}
    is locally Lipschitz continuous, and
    \item \label{cor:gradient:comp:lip:item3}
    it holds for all $i \in \cu{1, 2, \ldots, \width}$ that
        \begin{equation}
          \cu[\big]{\vartheta \in \R^\fd \colon i \notin \bfD^\vartheta } \ni \theta \mapsto \cG_{\width d + i } ( \theta ) \in \R
    \end{equation}
    is locally Lipschitz continuous.
\end{enumerate}
\end{cor}
\begin{cproof}{cor:gradient:comp:lip}
\Nobs that \cref{eq:loss:gradient} and \cref{cor:derivative:integral:lip} establish \cref{cor:gradient:comp:lip:item1,cor:gradient:comp:lip:item2,cor:gradient:comp:lip:item3}.
\end{cproof}

\subsection{Subdifferentials}
\label{subsection:subdiff}

\begin{definition}[Subdifferential]
\label{def:subdifferential}
Let $n \in \N$, $f \in C(\R^n, \R)$, $x \in \R^n$.
Then we denote by $\hat{\partial} f(x) \subseteq \R^n$ the set given by
    \begin{equation}
        \hat{\partial} f ( x ) = \cu*{ y \in \R^n \colon \liminf_{\R^n \backslash \cu{  0 } \ni h \to 0 } \rbr*{ \frac{f(x + h ) - f ( x ) - \spro{y , h } }{\norm{h}} } \geq 0  }.
    \end{equation}
\end{definition}

\cfclear
\begin{definition} [Limiting subdifferential] \label{def:limit:subdiff}
	Let $n \in \N$, $f \in C(\R^n, \R)$, $x \in \R^n$.
	Then we denote by $\partial f(x) \subseteq \R^n $ the set given by
	\begin{equation} \label{def:limit:subdiff:eq} \cfadd{def:subdifferential}
	\partial f ( x ) =
	 \textstyle\bigcap_{\varepsilon \in (0, \infty ) } \overline{\br*{\textstyle\bigcup\nolimits_{y \in \cu{z \in \R^n \colon \norm{x-z} < \varepsilon }} \hat{\partial} f ( y )  } }
	\end{equation}
	\cfload.
\end{definition}

\cfclear
\begin{lemma} \label{lem:subdiff:characterization}
	Let $n \in \N$, $f \in C(\R^n, \R)$, $x \in \R^n$. Then
	\begin{multline} \label{lem:subdiff:eq} \cfadd{def:subdifferential} \cfadd{def:limit:subdiff}
	\partial f ( x ) = \bigl\{ y \in \R^n \colon \exists \, z = (z_1, z_2) \colon \N \to \R^n \times \R^n \colon \bigl( \br[\big]{\forall \, k \in \N \colon z_2 ( k ) \in \hat{\partial} f(z_1(k))} , \\ \br[\big]{\limsup\nolimits_{k \to \infty} ( \norm{z_1(k) - x } + \norm{z_2(k) - y } ) = 0} \bigr) \bigr\}
	\end{multline}
	\cfload.
\end{lemma}
\begin{cproof}{lem:subdiff:characterization}
	\Nobs that \cref{def:limit:subdiff:eq} establishes \cref{lem:subdiff:eq}.
\end{cproof}

\cfclear

\begin{lemma} \label{lem:subdifferential:c1} \cfadd{def:subdifferential} \cfadd{def:limit:subdiff}
Let $n \in \N$,
$f \in C ( \R^n , \R)$, let $U \subseteq \R^n$ be open, assume $f | _U \in C^1 ( U , \R)$, and let $x \in U$. Then $\hat{\partial} f(x) = \partial f(x) = \cu{  ( \nabla f ) ( x ) } $ \cfload.
\end{lemma}
\begin{cproof}{lem:subdifferential:c1}
This is a direct consequence of, e.g., Rockafellar \& Wets~\cite[Exercise 8.8]{RockafellarWets1998}.
\end{cproof}

\cfclear
\begin{prop} \label{prop:loss:gradient:subdiff}
Assume \cref{setting:snn} and let $\theta \in \R^\fd$. Then $\cG ( \theta ) \in \partial \cL ( \theta )$ \cfadd{def:limit:subdiff}\cfload.
\end{prop}
\begin{cproof}{prop:loss:gradient:subdiff}
Throughout this proof let 
$\vartheta = (\vartheta_n)_{n \in \N} \colon \N \to \R^\fd$ satisfy for all $n \in \N$, $i \in \cu{ 1, 2, \ldots, \width}$, $j \in \cu{ 1, 2, \ldots, d}$ that $\w{\vartheta_n}_{i , j} = \w{\theta}_{i , j}$, $\b{\vartheta_n}_i = \b{\theta}_i - \frac{1}{n} \indicator{ \bfD^\theta } ( i )$, $\v{\vartheta_n} _ i = \v{\theta}_i$, and $\c{\vartheta_n} = \c{\theta}$.
We prove \cref{prop:loss:gradient:subdiff} through an application of \cref{lem:subdiff:characterization}.
\Nobs that for all $n \in \N$, $i \in \cu{ 1, 2, \ldots, \width } \backslash \bfD^\theta$ it holds that $\b{\vartheta_n}_i = \b{\theta}_i$. 
This implies for all $n \in \N$, $i \in \cu{ 1, 2, \ldots, \width } \backslash \bfD^\theta$ that
\begin{equation} \label{prop:loss:subdiff:eq1}
    i \notin \bfD^{\vartheta_n}.
\end{equation}
In addition, \nobs that for all $n \in \N$, $i \in \bfD^\theta$ 
it holds that $\b{\vartheta_n}_i = - \frac{1}{n} < 0$. This shows for all $n \in \N$, $i \in \bfD^\theta$ that
\begin{equation} \label{prop:loss:subdiff:eq2}
    i \notin \bfD^{\vartheta_n}.
\end{equation}
Hence, we obtain for all $n \in \N$ that $\bfD^{\vartheta_n} = \emptyset$. Combining this with \cref{prop:loss:continuously:diff} and \cref{lem:subdifferential:c1} demonstrates that for all $n \in \N$ it holds that $\hat{\partial} \cL ( \vartheta_n ) = \cu{  ( \nabla \cL ) ( \vartheta_n ) } = \cu{  \cG ( \vartheta_n ) }$ \cfadd{def:subdifferential}\cfload. Moreover, \nobs that $\lim_{n \to \infty} \vartheta_n = \theta$.
It thus remains to show that $ \cG ( \vartheta _n )$, $n \in \N$, converges to $\cG ( \theta )$.
\Nobs that
\cref{cor:gradient:comp:lip} ensures that for all $k \in \N \cap ( \width  d + \width , \fd]$ it holds that 
\begin{equation} \label{gradient:subdiff:eq1}
    \lim\nolimits_{n \to \infty} \cG_{k } ( \vartheta _n ) = \cG _ {k } ( \theta ).
\end{equation}
Furthermore, \nobs that \cref{cor:gradient:comp:lip},
\cref{prop:loss:subdiff:eq1},
and \cref{prop:loss:subdiff:eq2} assure that for all $i \in \cu{ 1, 2, \ldots, \width} \backslash \bfD^\theta$, $j \in \cu{ 1, 2, \ldots, d }$ it holds that 
\begin{equation} \label{gradient:subdiff:eq2}
    \lim\nolimits_{n \to \infty} \cG_{(i - 1 ) d + j} ( \vartheta _n ) = \cG _ {(i - 1 ) d + j } ( \theta ) \qqandqq \lim\nolimits_{n \to \infty} \cG_{\width d + i } ( \vartheta_n ) = \cG_{\width d + i } ( \theta ).
\end{equation}
In addition, \nobs that for all $n \in \N$, $i \in \bfD^\theta$
we have that $I_i^{\vartheta_n } = I_i^\theta = \emptyset$. Hence, we obtain for all $i \in \bfD^\theta$, $j \in \cu{ 1, 2, \ldots, d}$ 
that 
\begin{equation} \label{gradient:subdiff:eq3}
    \lim\nolimits_{n \to \infty} \cG_{(i - 1 ) d + j} ( \vartheta_n ) = 0 = \cG_{(i - 1 ) d + j} ( \theta ) \qqandqq \lim\nolimits_{n \to \infty} \cG _{\width d + i } ( \vartheta_n ) = 0 = \cG_{\width d + i } ( \theta ).
\end{equation}
Combining this, \cref{gradient:subdiff:eq1}, and \cref{gradient:subdiff:eq2} demonstrates that $\lim_{n \to \infty} \cG ( \vartheta _ n ) = \cG ( \theta )$.
 This and \cref{lem:subdiff:characterization} assure that $\cG ( \theta ) \in \partial \cL ( \theta )$. 
\end{cproof}

\section{Existence and uniqueness properties for solutions of gradient flows (GFs)}
\label{section:gf:existence}

In this section we employ the local Lipschitz continuity result 
for the generalized gradient function in \cref{cor:gradient:comp:lip} 
from \cref{section:risk:diff}
to establish existence and uniqueness results for solutions of GF differential equations. 
Specifically, in \cref{prop:gf:existence} in \cref{subsection:existence} below 
we prove the existence of solutions GF differential equations, 
in \cref{lem:gf:unique} in \cref{subsection:uniqueness} below 
we establish the uniqueness of solutions of GF differential equations 
among a suitable class of GF solutions, 
and in \cref{theo:gf:exist:unique} in \cref{subsection:exist:unique} below 
we combine \cref{prop:gf:existence,lem:gf:unique} to establish 
the unique existence of solutions of GF differential equations 
among a suitable class of GF solutions. \cref{theo:intro:existence} in the introduction is an immediate consequence of \cref{theo:gf:exist:unique}.

Roughly speaking, we show in \cref{theo:gf:exist:unique}
the unique existence of solutions of GF differential equations among the 
class of GF solutions which satisfy that the set 
of all degenerate neurons of the GF solution at time $ t \in [0,\infty) $ 
is non-decreasing in the time variable $ t \in [0,\infty) $. 
In other words, in \cref{theo:gf:exist:unique} we prove the 
unique existence of GF solutions 
with the property that once a neuron has become degenerate 
it will remain degenerate for subsequent times.

Our strategy of the proof of \cref{theo:gf:exist:unique} and \cref{prop:gf:existence}, 
respectively, can, loosely speaking, be described as follows.
\cref{cor:gradient:comp:lip} above implies that the components 
of the generalized gradient function 
$ \cG \colon \R^{ \fd } \to \R^{ \fd } $ 
corresponding to non-degenerate neurons are locally Lipschitz continuous 
so that the classical Picard-Lindel\"{o}f local existence and uniqueness theorem 
for ordinary differential equations can be brought into play for those components. 
On the other hand, if at some time $ t \in [0, \infty ) $ the $i$-th neuron 
is degenerate, 
then 
\cref{prop:loss:approximate:gradient} above
shows that the corresponding components 
of the generalized gradient function 
$ \cG \colon \R^{ \fd } \to \R^{ \fd } $ 
vanish. 
The GF differential equation is thus satisfied 
if the neuron remains degenerate at all subsequent times $ s \in [t,\infty) $. 
Using these arguments we prove 
in \cref{prop:gf:existence}
the existence of GF solutions 
by induction on the number of non-degenerate neurons of the initial value.

\subsection{Existence properties for solutions of GF differential equations}
\label{subsection:existence}

\begin{prop} \label{prop:gf:existence}
Assume \cref{setting:snn} and let $\theta \in \R^\fd$. Then there exists $\Theta \in C([0, \infty ) , \R^\fd)$ which satisfies for all $t \in [0, \infty)$, $s \in [0, \infty )$ that 
\begin{equation} \label{eq:gf:conditions}
    \Theta_t = \theta - \int_0^t \cG ( \Theta_u ) \, \d u \qqandqq \bfD^{\Theta_t} \subseteq \bfD^{\Theta_s}.
\end{equation}
\end{prop}
\begin{cproof}{prop:gf:existence}
We prove the statement by induction on the quantity $\width - \# ( \bfD^\theta ) \in \N \cap [0, \width ]$.
Assume first that $\width - \# ( \bfD^\theta ) = 0$, i.e.,~$\bfD^\theta = \cu{1, 2, \ldots, \width}$. \Nobs that this implies that $\w{\theta} = 0$ and $\b{\theta} = 0$.
In the following let $\kappa \in \R$ satisfy 
\begin{equation}
    \kappa = \int_{ [ \scra , \scrb ] ^d}  f ( x ) \dens ( x ) \, \lambda ( \d x ).
\end{equation}
\Nobs that the Picard--Lindelöf Theorem shows that there exists a unique $c \in C([0, \infty ) , \R)$ which satisfies for all $t \in [0, \infty )$ that
\begin{equation} \label{eq:gf:induct:start:defc}
    c(0) = \c{\theta} \qquad \text{and} \qquad c(t) = c(0) + 2 \kappa t - 2 \rbr*{\int_{ [\scra , \scrb ] ^d } \dens ( x ) \, \lambda ( \d x ) } \rbr*{ \int_0^t c(s) \, \d s }.
\end{equation}
Next let $\Theta \in C([0, \infty ) , \R^\fd) $ satisfy for all $t \in [0, \infty)$, $i \in \cu{ 1, 2, \ldots, \width }$, $j \in \cu{1, 2, \ldots, d}$ that
\begin{equation} \label{eq:gf:induct:start:deftheta}
    \w{\Theta_t}_{i , j} = \w{\theta}_{i,j} = \b{\Theta_t}_i = \b{\theta}_i = 0, \qquad \v{\Theta_t}_i = \v{\theta}_i, \qquad \text{and} \qquad \c{\Theta_t}  = c ( t ) .
\end{equation}
\Nobs that \cref{eq:gf:induct:start:defc,eq:gf:induct:start:deftheta,eq:loss:gradient} ensure for all $t \in [0, \infty)$ that
\begin{equation} \label{eq:gf:induct:start:intc}
\begin{split}
 \c{\Theta_t } &= \c{\theta} + 2 \kappa t - 2 \rbr*{\int_{ [\scra , \scrb ] ^d } \dens ( x ) \, \lambda ( \d x ) } \rbr*{ \int_0^t \c{\Theta_s } \, \d s } \\
 &= \c{\theta} - 2 \int_0^t \rbr*{ - \kappa + \int_{ [ \scra , \scrb ] ^d } \c{\Theta_s} \dens ( x ) \, \lambda ( \d x ) } \, \d s \\
&= \c{\theta} - 2 \int_0^t \int_{ [ \scra , \scrb ]^d } \rbr*{\c{\Theta_s} + \smallsum_{i=1}^\width \br[\big]{ \v{\Theta_s}_i \max \cu[\big]{ \b{\Theta_s}_i + \smallsum_{j=1}^d \w{\Theta_s }_{i , j} x_j , 0 } } - f ( x ) } \dens ( x ) \, \lambda ( \d x ) \, \d s\\
&= \c{\theta} - 2\int_0^t \int_{ [ \scra , \scrb ] ^d } ( \realization{\Theta_s} ( x ) - f ( x ) ) \dens ( x ) \, \lambda ( \d x ) \, \d s = \c{\theta} - \int_0^t \cG_\fd ( \Theta_s ) \, \d s. 
\end{split}
\end{equation}
Next \nobs that \cref{eq:gf:induct:start:deftheta} and \cref{eq:loss:gradient} show for all $t \in [0, \infty)$, $i \in \N \cap [1, \fd)$ that $\bfD^{\Theta_t} = \cu{1, 2, \ldots, \width}$ and
$\cG_i ( \Theta_t ) = 0$. Combining this with \cref{eq:gf:induct:start:deftheta} and \cref{eq:gf:induct:start:intc} proves that $\Theta$ satisfies \cref{eq:gf:conditions}.
This establishes the claim in the case $\# ( \bfD^\theta ) = \width$.

For the induction step assume that $\# ( \bfD^\theta ) < \width$ and assume that for all $\vartheta \in \R^\fd$ with $\# ( \bfD^\vartheta ) > \# ( \bfD^\theta)$ there exists $\Theta \in C([0, \infty ) , \R^\fd)$ which satisfies for all $t \in [0, \infty)$, $s \in [0, \infty )$ that 
    $\Theta_t = \vartheta - \int_0^t \cG ( \Theta_u ) \, \d u$ and $\bfD^{\Theta_t} \subseteq \bfD^{\Theta_s}$.
In the following let $U \subseteq \R^\fd$ satisfy 
\begin{equation} \label{prop:gf:existence:eq:defu}
    U = \cu[\big]{\vartheta \in \R^\fd \colon \bfD^\vartheta \subseteq \bfD^\theta }
\end{equation}
and let $\fG \colon U \to \R^\fd$ satisfy for all $\vartheta \in U$, $i \in \cu{ 1, 2, \ldots, \fd }$ that
\begin{equation} \label{prop:gf:existence:eqfg}
    \fG _ i ( \vartheta ) = \begin{cases}
     0 & \colon i \in \cu{(\ell - 1 ) d + j \colon \ell \in \bfD^\theta , j \in \N \cap [1 , d] } \cup \cu{\width d + \ell \colon \ell \in \bfD^\theta } \\
     \cG_i ( \vartheta ) & \colon \text{else}.
    \end{cases}
\end{equation}
\Nobs that \cref{prop:gf:existence:eq:defu} assures that
$U \subseteq \R^\fd$ is open.
In addition, \nobs that \cref{cor:gradient:comp:lip} implies that $\fG$ is locally Lipschitz continuous.
Combining this with the Picard--Lindelöf Theorem demonstrates
that there exist a unique maximal $\tau \in (0, \infty]$ and $\Psi \in C([0, \tau ), U)$ which satisfy for all $t \in [0, \tau)$ that
\begin{equation} \label{eq:gf:exist:defpsi}
    \Psi_t = \theta - \int_0^t \fG ( \Psi_u ) \, \d u .
\end{equation}
Next \nobs that the fact that for all $\vartheta \in U$, $i \in \cu{(\ell - 1 ) d + j \colon \ell \in \bfD^\theta , j \in \N \cap [1 , d] } \cup \cu{\width d + \ell \colon \ell \in \bfD^\theta }$ it holds that $\fG_i ( \vartheta ) = 0$ ensures that for all $t \in [0, \tau)$, $i \in \bfD^\theta$, $j \in \cu{1, 2, \ldots, d}$ we have that
\begin{equation} \label{eq:gf:exist:degen}
    \w{\Psi_t}_{i , j} = \w{\theta}_{i , j} = \b{\Psi_t}_i = \b{\theta}_i = 0 \qqandqq \v{\Psi_t}_i = \v{\theta}_i.
\end{equation}
This, \cref{prop:gf:existence:eqfg},
and \cref{eq:loss:gradient} demonstrate for all $t \in [0, \tau)$ that $\cG ( \Psi_t ) = \fG ( \Psi_t )$. 
In addition, \nobs that \cref{prop:gf:existence:eq:defu} and
\cref{eq:gf:exist:degen} imply for all $t \in [0, \tau)$ that $\bfD^{\Psi_t} = \bfD^\theta$.
Hence, if $\tau = \infty$ then $\Psi$ satisfies \cref{eq:gf:conditions}.
Next assume that $\tau < \infty$. 
\Nobs that the Cauchy-Schwarz inequality and \cite[Lemma 3.1]{JentzenRiekertFlow} prove for all $s,t \in [0, \tau)$ with $s \leq t$ that
\begin{equation}
\begin{split}
    \norm{\Psi_t - \Psi_s} &\leq \int_s^t \norm{\cG ( \Psi_u ) } \, \d u \leq (t-s)^{1/2} \br*{\int_s^t \norm{\cG ( \Psi_u ) } ^2 \, \d u }^{1/2} \\
     &\leq  (t-s)^{1/2} \br*{\int_0^t \norm{\cG ( \Psi_u ) } ^2 \, \d u }^{1/2} 
     = (t - s )^{1/2} \rbr[\big]{\cL ( \Psi_0 ) - \cL ( \Psi_t )}^{1/2} \\
     &\leq  (t - s )^{1/2} \rbr[\big]{\cL ( \Psi_0 ) }^{1/2} .
\end{split}
\end{equation}
Hence, we obtain for all $(t_n) _{n \in \N} \subseteq [0, \tau)$ with $\liminf_{n \to \infty} t_n = \tau$ that $(\Psi_{t_n})$ is a Cauchy sequence. This implies that $\vartheta := \lim_{t \uparrow \tau} \Psi_t \in \R^\fd$ exists.
Furthermore, \nobs that the fact that $\tau$ is maximal proves that $\vartheta \notin U$.
Therefore, we have that $\bfD^\vartheta \backslash \bfD^\theta \not= \emptyset$.
Moreover, \nobs that \cref{eq:gf:exist:degen} shows that for all $i \in \bfD^\theta $, $j \in \cu{1, 2, \ldots, d}$ it holds that $\w{\vartheta}_{i , j} = \b{\vartheta}_i = 0$ and, therefore, $i \in \bfD^\vartheta$. 
This demonstrates that $\# ( \bfD^\vartheta ) > \# (\bfD^\theta)$.
Combining this with the induction hypothesis ensures that there exists $\Phi \in C([0, \infty), \R^\fd)$ which satisfies for all $t \in [0, \infty)$, $s \in [0 , \infty )$ that
\begin{equation}
    \Phi_t = \vartheta - \int_0^t \cG ( \Phi_u ) \, \d u \qqandqq \bfD^{\Phi_t} \subseteq \bfD^{\Phi_s} .
\end{equation}
In the following let $\Theta \colon [0, \infty ) \to \R^\fd$ satisfy for all $t \in [0, \infty)$ that
\begin{equation}
    \Theta_t = \begin{cases}
    \Psi_t & \colon t \in [0, \tau) \\
    \Phi_{t - \tau} & \colon t \in [\tau , \infty ).
    \end{cases}
\end{equation}
\Nobs that the fact that $\vartheta = \lim_{t \uparrow \tau} \Psi_t$ and the fact that $\Phi_0 = \vartheta$ imply that $\Theta$ is continuous. Furthermore, \nobs that the fact that $\cG$ is locally bounded and \cref{eq:gf:exist:defpsi} ensure that
\begin{equation}
    \Theta_\tau = \vartheta = \lim_{t \uparrow \tau} \Psi_t = \lim_{t \uparrow \tau} \br*{ \theta - \int_0^t \cG ( \Psi_s ) \, \d s } = \theta - \int_0^\tau \cG ( \Psi_s ) \, \d s = \theta - \int_0^\tau \cG ( \Theta_s ) \, \d s.
\end{equation}
Hence, we obtain for all $t \in [\tau, \infty)$ that
\begin{equation}
    \begin{split}
        \Theta_t &= (\Theta_t - \Theta_\tau ) + \Theta_\tau = (\Phi_{t - \tau} - \Phi_0 ) + \Theta_\tau = - \int_0^{t - \tau} \cG ( \Phi_s ) \, \d s + \theta - \int_0^\tau \cG ( \Theta_s ) \, \d s \\
        &= - \int_t^\tau \cG ( \Theta_s ) + \theta - \int_0^\tau \cG ( \Theta_s ) \, \d s = \theta - \int_0^t \cG ( \Theta_s ) \, \d s.
    \end{split}
\end{equation}
This shows that $\Theta$ satisfies \cref{eq:gf:conditions}.
\end{cproof}

\subsection{Uniqueness properties for solutions of GF differential equations}
\label{subsection:uniqueness}

\begin{lemma} \label{lem:gf:unique}
Assume \cref{setting:snn}
and  let $\theta \in \R^\fd$, $\Theta^1, \Theta^2 \in C([0, \infty ) , \R^\fd)$ satisfy for all $t \in [0, \infty)$, $s \in [t, \infty)$, $k \in \cu{ 1, 2}$ that
\begin{equation} \label{eq:gf:unique}
    \Theta_t^k = \theta - \int_0^t \cG ( \Theta_u^k ) \, \d u \qqandqq \bfD^{\Theta_t^k } \subseteq \bfD^{\Theta_s^k }.
\end{equation}
Then it holds for all $t \in [0, \infty)$ that $\Theta_t^1 = \Theta_t^2$.
\end{lemma}
\begin{cproof}{lem:gf:unique}
Assume for the sake of contradiction that there exists $t \in [0, \infty)$ such that $\Theta_t^1 \not= \Theta_t^2$.
By translating the variable $t$ if necessary, we may assume without loss of generality that $\inf \cu*{t \in [0, \infty) \colon \Theta_t^1 \not= \Theta_t^2} = 0$.
Next \nobs that the fact that $\Theta^1$ and $\Theta^2$ are continuous implies that there exists $\delta \in (0, \infty)$ which satisfies for all $t \in [0, \delta]$, $k \in \cu{ 1, 2}$ that $\bfD^{\Theta_t^k} \subseteq \bfD^\theta$.
Furthermore, \nobs that \cref{eq:gf:unique} ensures for all $t \in [0, \infty)$, $i \in  \bfD^\theta$, $k \in \cu{ 1, 2}$ that $i \in \bfD^{\Theta_t^k}$.
Hence, we obtain for all $t \in [0, \infty)$, $i \in  \bfD^\theta$, $j \in \cu{1, 2, \ldots, d}$, $k \in \cu{ 1, 2}$ that 
\begin{equation} \label{eq:gf:unique:degen}
    \cG_{(i - 1 ) d + j } ( \Theta_t^k ) = \cG_{\width d + i } ( \Theta_t^k ) = \cG_{\width ( d+1 ) + i } ( \Theta _t^k ) = 0 .
\end{equation}
In addition, \nobs that the fact that $\Theta^1$ and $\Theta^2$ are continuous implies that there exists a compact $K \subseteq \cu{\vartheta \in \R^\fd \colon \bfD^\vartheta \subseteq \bfD^\theta }$ which satisfies for all $t \in [0, \delta]$, $k \in \cu{1, 2}$ that $\Theta_t^k \in K$. 
Moreover, \nobs that \cref{cor:gradient:comp:lip} proves that for all $i \in \cu{1, 2, \ldots, \width} \backslash \bfD^\theta$, $j \in \cu{1, 2, \ldots, d}$ it holds that $\cG_{(i - 1 ) d + j } , \cG_{\width d + i }, \cG_{\width ( d+1 ) + i } , \cG _\fd \colon K \to \R$ are Lipschitz continuous.
This and \cref{eq:gf:unique:degen} show that there exists $L \in (0, \infty)$ such that for all $t \in [0, \delta]$ we have that 
\begin{equation} \label{eq:gf:unique:lipest}
    \norm{\cG ( \Theta_t^1 ) - \cG ( \Theta_t^2 ) } \leq L \norm{\Theta_t^1 - \Theta_t^2}.
\end{equation}
In the following let $M \colon [0, \infty) \to [0, \infty)$ satisfy for all $t \in [0, \infty)$ that $M_t = \sup_{s \in (0,t] } \norm{\Theta_s^1 - \Theta_s^2}$. \Nobs that the fact that $\inf \cu*{t \in [0, \infty) \colon \Theta_t^1 \not= \Theta_t^2} = 0$ proves for all $t \in (0, \infty)$ that $M_t > 0$. Moreover, \nobs that \cref{eq:gf:unique:lipest} ensures for all $t \in (0, \delta )$ that
\begin{equation}
    \begin{split}
        \norm{\Theta_t^1 - \Theta_t^2} 
        &= \norm*{ \int_0^t \cG ( \Theta_u^1) \, \d u - \int_0^t \cG ( \Theta_u^2 ) \, \d u } \leq \int_0^t \norm{\cG ( \Theta_u^1) - \cG ( \Theta_u^2 ) } \, \d u \\
        &\leq L \int_0^t \norm{\Theta_u^1 - \Theta_u^2} \, \d u \leq L t M_t.
    \end{split}
\end{equation}
Combining this with the fact that $M$ is non-decreasing shows for all $t \in (0, \delta)$, $s \in (0, t]$ that
\begin{equation}
    \norm{\Theta_s^1 - \Theta_s^2} \leq L s M_s \leq L t M_t.
\end{equation}
This demonstrates for all $t \in (0, \min \cu{L^{-1}, \delta } ) $ that
\begin{equation}
    0 < M_t \leq Lt M_t < M_t,
\end{equation}
which is a contradiction.
\end{cproof}

\subsection{Existence and uniqueness properties for solutions of GF differential equations}
\label{subsection:exist:unique}

\begin{theorem} \label{theo:gf:exist:unique}
Assume \cref{setting:snn} and let $\theta \in \R^\fd$.
Then there exists a unique $\Theta \in C([0, \infty ) , \R^\fd)$ which satisfies for all $t \in [0, \infty)$, $s \in [t, \infty)$ that
\begin{equation}
    \Theta_t = \theta - \int_0^t \cG ( \Theta_u ) \, \d u \qqandqq \bfD^{\Theta_t} \subseteq \bfD^{\Theta_s }.
\end{equation}
\end{theorem}
\begin{cproof}{theo:gf:exist:unique}
\cref{prop:gf:existence} establishes the existence and 
\cref{lem:gf:unique} establishes the uniqueness. 
\end{cproof}

\section{Semialgebraic sets and functions}
\label{section:semialgebraic}

In this section we establish in \cref{cor:loss:semialgebraic} in \cref{subsection:risk:semialg} below that under the assumption that both the target function $f \colon [ \scra , \scrb ] ^d \to \R$ and the unnormalized density function $\dens \colon [\scra , \scrb ]^d \to [0,\infty)$ are piecewise polynomial in the sense of \cref{def:multidim:piece:polyn} in \cref{subsection:risk:semialg} we have that the risk function $\cL \colon \R^{ \fd } \to \R$ 
is a semialgebraic function in the sense of \cref{def:semialgebraic:function} in \cref{subsection:semialgebraic:def}. In \cref{def:multidim:piece:polyn} we specify precisely what we mean by a piecewise polynomial function, in \cref{def:semialgebraic:set} in \cref{subsection:semialgebraic:def} we recall the notion of a 
semialgebraic set, and in \cref{def:semialgebraic:function} we recall the notion of a 
semialgebraic function.
 In the scientific literature \cref{def:semialgebraic:set,def:semialgebraic:function} can in a slightly different presentational form, e.g., be found in Bierstone \& Milman~\cite[Definitions 1.1 and 1.2]{BierstoneMilman1988} and Attouch et al.~\cite[Definition 2.1]{AttouchBolteSvaiter2013}. 

Note that the risk function $ \cL \colon \R^{ \fd } \to \R $ is given through 
a parametric integral in the sense that for all $ \theta \in \R^{ \fd } $ 
we have that 
$ \cL( \theta ) = \int_{[ \scra , \scrb ] ^d} ( f ( y ) - \realization{\theta} (y) )^2 \dens ( y ) \, \lambda ( \d y ) $. In general, parametric integrals of semialgebraic functions are no longer semialgebraic functions and the characterization of functions that can occur as such integrals is quite involved (cf.~Kaiser~\cite{Kaiser2013}).
This is the reason why we introduce in \cref{def:function:amn} in \cref{subsection:semialg:integrals} below a suitable subclass of the class of semialgebraic functions which is rich enough to contain the realization functions of ANNs with ReLU activation (cf.~\cref{cor:loss:semialgebraic:eq:real:amn} in \cref{subsection:semialg:integrals} below) and which can be shown to be closed under integration (cf.~\cref{prop:integrals:amn} in \cref{subsection:semialg:integrals} below for the precise statement).

\subsection{Semialgebraic sets and functions}
\label{subsection:semialgebraic:def}

\begin{definition}[Set of polynomials]
	\label{def:polynomial}	
	Let $n \in \N_0$. Then we denote by $\polyn_n \subseteq C(\R^n , \R)$ 
	the set\footnote{Note that $ \R^0 = \{ 0 \} $, 
	$ C( \R^0, \R ) = C( \{ 0 \}, \R ) $, and 
	$ \#( C( \R^0, \R ) ) = \#( C( \{ 0 \}, \R ) ) = \infty $. In particular, this shows 
	for all $ n \in \N_0 $ that $ \operatorname{dim}( \R^n ) = n $ 
	and $ \#( C( \R^n, \R ) ) = \infty $.} 
	of all polynomials from $\R^n$ to $\R$.
\end{definition}

\cfclear
\begin{definition}[Semialgebraic sets]
	 \label{def:semialgebraic:set}
Let $n \in \N$ and let $A \subseteq \R^n$ be a set. 
    Then we say that $A$ is a semialgebraic set if and only if there exist $k \in \N$,
     $(P_{i,j, \ell })_{ (i, j, \ell ) \in \cu{1, 2, \ldots, k} ^2 \times \cu{0,1}} \subseteq \polyn_n $ such that
    \begin{equation} \cfadd{def:polynomial}
        A = \bigcup_{i=1}^k \bigcap_{j=1}^k \cu*{ x \in \R^n \colon P_{i, j, 0} ( x ) = 0 < P_{i , j , 1} ( x )  }
    \end{equation}
    \cfload.
\end{definition}

\cfclear
\begin{definition}[Semialgebraic functions]
	 \label{def:semialgebraic:function}
 Let $m , n \in \N$ and let $f \colon \R^n \to \R^m$ be a function. Then we say that $f$ is a semialgebraic function if and only if it holds that $\cu{ ( x , f ( x ) ) \colon x \in \R^n } \subseteq \R^{m+n}$ is a semialgebraic set \cfadd{def:semialgebraic:set}\cfload.
\end{definition}

\cfclear
\begin{lemma} \label{lem:sum:semialgebraic}
Let $n \in \N$ and let $f,g \colon \R^n \to \R$ be semialgebraic functions \cfadd{def:semialgebraic:function}\cfload.
Then
\begin{enumerate} [ label = (\roman*)]
    \item \label{lem:sum:semialgebraic:item1} it holds that $\R^n \ni x \mapsto f(x) + g(x) \in \R$ is semialgebraic and
    \item \label{lem:sum:semialgebraic:item2} it holds that $\R^n \ni x \mapsto f(x)  g(x) \in \R$ is semialgebraic.
\end{enumerate}
\end{lemma}
\begin{cproof}{lem:sum:semialgebraic}
	\Nobs that, e.g., Coste~\cite[Corollary 2.9]{Coste2000} (see, e.g., also Bierstone \& Milman~\cite[Section 1]{BierstoneMilman1988}) establishes \cref{lem:sum:semialgebraic:item1,lem:sum:semialgebraic:item2}.
\end{cproof}

\subsection{On the semialgebraic property of certain parametric integrals}
\label{subsection:semialg:integrals}

\cfclear
\begin{definition} [Set of rational functions]
	\label{def:rational:function}
Let $n \in \N$. 
Then we denote by $\ratio_n$ the set given by
\begin{equation} \cfadd{def:polynomial}
   \ratio_n = \cu*{R \colon \R^n \to \R \colon \br*{ \exists \, P, Q \in \polyn_n \colon \forall \, x \in \R^n \colon  R(x) = \begin{cases}
     \frac{P(x) }{ Q ( x ) } & \colon Q ( x ) \not= 0 \\[0.5ex]
     0 & \colon Q ( x ) = 0 
     \end{cases} }}
\end{equation}
\cfload.
\end{definition}

\cfclear
\begin{definition} \label{def:function:amn}
Let $m \in \N$, $n \in \N_0$. Then we denote by $\scrA_{m,n}$ the $\R$-vector space given by
\begin{multline} \cfadd{def:polynomial}\cfadd{def:rational:function}
  \scrA_{m,n} = \operatorname{span} \Bigl( \Bigl\{ f \colon \R^m \times \R^n \to \R \colon \Bigl[  
 \exists \, r \in \N, \,
   A_1, A_2, \ldots, A_r \in \cu{ \cu{0}, [0, \infty ), (0, \infty )}, \\
   R \in \ratio_m , \,
    Q \in \polyn_n, 
   \,
    P = (P_{i,j})_{ (i,j) \in \cu{1, 2, \ldots, r } \times \cu{0, 1, \ldots, n }} \subseteq \polyn_m \colon 
    \forall \, \theta \in \R^m , \, x = (x_1, \ldots, x_n) \in \R^n \colon \\
    	f ( \theta , x ) = R ( \theta ) Q ( x ) \br[\big]{ \textstyle\prod_{i=1}^r  \indicator{A_i} \rbr[\big]{ P_{i,0} ( \theta ) + \smallsum_{j = 1}^n P_{i,j} ( \theta ) x_j } } \Bigr] \Bigr\} \Bigr)
\end{multline}
\cfload.
\end{definition}

\cfclear

\begin{lemma} \label{lem:amn:semialgebraic}
Let $m \in \N$, $f \in \scrA_{m, 0 }$ \cfadd{def:function:amn}\cfload. Then $f$ is semialgebraic \cfadd{def:semialgebraic:function}\cfload.
\end{lemma}

\begin{cproof}{lem:amn:semialgebraic}
Throughout this proof let $r \in \N$,
$A_1, A_2, \ldots, A_r \in \cu{ \cu{0}, [0, \infty ), (0, \infty )}$,
$R \in \ratio_m$,
$P = (P_i)_{ i \in \cu{1, 2, \ldots, r }} \subseteq \polyn_m$,
and let $g \colon \R^m  \to \R$ satisfy for all $ \theta \in \R^m $ that
\begin{equation} \label{lem:amn:semialg:eq:defg} \cfadd{def:rational:function} \cfadd{def:polynomial}
    g(\theta) = R ( \theta ) \textstyle\prod_{i= 1}^r \indicator{A_i} \rbr{ P_i ( \theta ) }
\end{equation}
\cfload.
Since sums of semialgebraic functions are again semialgebraic (cf.~\cref{lem:sum:semialgebraic}), it suffices to show that $g$ is semialgebraic. Furthermore, \nobs that for all $y \in \R$ it holds that $\indicator{(0, \infty )} ( y ) = 1 - \indicator{[0, \infty ) } ( - y )$ and $\indicator{\cu{0}} ( y ) = \indicator{[0, \infty ) } ( y ) \indicator{[0, \infty ) } ( - y )$. Hence, by linearity we may assume for all $i \in \cu{1, 2, \ldots, r }$ that $A_i = [0, \infty ) $.
Next let $Q_1, Q_2 \in \polyn_m$ satisfy for all $x \in \R^m$ that
\begin{equation}
     R(x) = \begin{cases}
      \frac{Q_1 ( x ) }{ Q_2 ( x ) } & \colon Q_2 ( x ) \not= 0 \\
      0 & \colon Q_2 ( x ) = 0.
      \end{cases}
 \end{equation}
\Nobs that the graph of $\R^m \ni \theta \mapsto R(\theta) \in \R$ is given by
\begin{multline}
    \cu*{(\theta , y ) \in \R^m \times \R \colon Q_2(\theta ) = 0 , \, y = 0 } \\
    \cup \cu{(\theta , y ) \in \R^m \times \R \colon Q_2 (\theta ) \not= 0 , \, Q_2 ( \theta ) y - Q_1 ( \theta ) = 0}.
\end{multline}
Since both of these sets are described by polynomial equations and inequalities,
it follows that $\R^m  \ni \theta \mapsto R(\theta) \in \R$ is semialgebraic.
In addition,
\nobs that for all $i \in \cu{1, 2, \ldots, r}$ the graph of $\R^m \ni \theta  \mapsto \indicator{[0, \infty ) } \rbr{ P_i ( \theta ) } \in \R$ is given by
\begin{equation}
    \cu*{(\theta , y ) \in \R^m \times \R \colon P_i ( \theta ) < 0 , \, y = 0 } 
    \cup  \cu*{(\theta , y ) \in \R^m \times \R \colon P_i ( \theta ) \geq 0 , \, y = 1 }.
\end{equation}
This demonstrates for all $i \in \cu{1, 2, \ldots, r}$ that $\R^m \ni \theta \mapsto \indicator{[0, \infty ) } \rbr{ P_i ( \theta ) } \in \R$ is semialgebraic. Combining this and \cref{lem:amn:semialg:eq:defg} with \cref{lem:sum:semialgebraic} demonstrates that $g$ is semialgebraic.
\end{cproof}

\cfclear
\begin{prop} \label{prop:integrals:amn}
Let $m,n \in \N$, $\scra \in \R$, $\scrb \in ( \scra , \infty)$, $f \in \scrA_{m,n}$ \cfadd{def:function:amn}\cfload. Then 
\begin{equation}
    \br*{ \R^m \times \R^{n-1} \ni (\theta, x_1, \ldots, x_{n-1} ) \mapsto \int_\scra^\scrb f ( \theta , x_1, \ldots, x_n ) \, \d x_n  \in \R } \in \scrA_{m, n-1}.
\end{equation}
\end{prop}

\begin{cproof}{prop:integrals:amn}
By linearity of the integral it suffices to consider a function $f $ of the form
\begin{equation}
    f(\theta , x ) = R ( \theta ) Q ( x ) \prod_{i = 1}^r \indicator{A_i} \rbr[\big]{ P_{i,0} ( \theta ) +  \smallsum_{j = 1}^n P_{i , j} ( \theta ) x_j  }
\end{equation}
where
$r \in \N$,
$\rbr{P_{i,j}}_{(i , j )  \in \cu{1, 2, \ldots, r} \times \cu{0, 1, \ldots, n } } \subseteq \polyn_m $, 
 $A_1, A_2, \ldots, A_r \in  \cu{ \cu{0} , (0, \infty ) , [0, \infty ) }$,
 $Q \in \polyn_n$, and
$R \in \ratio_m$ \cfadd{def:polynomial}\cfadd{def:rational:function}\cfload. 
Moreover, \nobs that for all $y \in \R$ it holds that $\indicator{(0, \infty )} ( y ) = 1 - \indicator{[0, \infty ) } ( - y )$ and $\indicator{\cu{0}} ( y ) = \indicator{[0, \infty ) } ( y ) \indicator{[0, \infty ) } ( - y )$. Hence, by linearity we may assume that $A_i = [0, \infty ) $ for all $i \in \cu{1, 2, \ldots, r }$.
Furthermore,
by linearity we may assume that $Q$ is of the form 
\begin{equation}
    Q(x_1, \ldots, x_n) = \textstyle\prod_{\ell=1}^n ( x_\ell ) ^{i_\ell} 
\end{equation}
with $i_1, i_2, \ldots, i_n \in \N_0$.
In the following let $\sign \colon \R \to \R$ satisfy for all $x \in \R$ that $\sign ( x ) = \indicator{(0, \infty)} ( x ) - \indicator{(0, \infty ) } ( - x )$,
for every $\theta \in \R^m$, $k \in \cu{-1, 0, 1}$ let $\cS_k^\theta \subseteq \cu{1, 2, \ldots, r}$ satisfy $\cS_k^\theta = \cu{ i \in \cu{1, 2, \ldots, r } \colon \sign ( P_{i , n} ( \theta ) ) = k }$,
and for every $i \in \cu{1, 2, \ldots, r}$ let $Z_i \colon \R^m \times \R^n \to \R$ satisfy for all $(\theta , x ) \in \R^m \times \R^n$ that
\begin{equation}
    Z_i ( \theta , x ) =  - P_{i,0} ( \theta ) - \smallsum_{j=1}^{n-1} P_{i,j} ( \theta ) x_j .
\end{equation}
\Nobs that for all $\theta \in \R^m$, $x = (x_1, \ldots , x_n) \in \R^n$ with $x_n \in [\scra , \scrb]$, $f(\theta,x)$ can only be nonzero if 
\begin{equation}
\begin{split}
    \forall \, i \in \cS^\theta_1 &\colon x_n \geq  \frac{Z_i ( \theta , x )}{ P_{i , n} ( \theta ) }, \\
    \forall \, i \in \cS^\theta_{-1} &\colon x_n \leq  \frac{Z_i ( \theta , x )}{ P_{i , n} ( \theta ) }, \\
    \forall \, i \in \cS^\theta_0 &\colon  - Z_i ( \theta , x ) \ge 0.
\end{split}
\end{equation}
Hence, if for given $\theta \in \R^m$, $(x_1, \ldots, x_{n-1} ) \in \R^{n-1}$ there exists $x_n \in [\scra , \scrb]$ which satisfies these conditions, we have
\begin{equation}
\begin{split}
   &  \int_\scra^\scrb f ( \theta , x_1, \ldots, x_n ) \, \d x_n 
   \\
   & = \frac{ R ( \theta )}{i_n + 1 } \rbr*{ \textstyle \prod_{\ell=1}^{n-1} x_\ell^{i_\ell} } 
   \br*{ \rbr*{ \min \cu*{\scrb , \min_{j \in \cS_{-1}^\theta}  \frac{Z_j ( \theta , x ) }{ P_{j , n} ( \theta ) } } }^{\! i_n + 1} \! \! - \rbr*{ \max \cu*{ \scra,  \max_{j \in \cS_1^\theta}  \frac{ Z_j ( \theta , x ) }{ P_{j , n} ( \theta ) } } }^{\! i_n + 1} }.
    \end{split}
\end{equation}
Otherwise, we have that $\int_\scra^\scrb f ( \theta , x_1, \ldots, x_n ) \, \d x_n  = 0$.
It remains to write these expressions in the different cases as a sum of functions of the required form by introducing suitable indicator functions. \Nobs that there are four possible cases where the integral is nonzero:
\begin{itemize}
    \item It holds that $\scra < \max_{j \in \cS_1^\theta}  \frac{ Z_j ( \theta , x ) }{ P_{j , n} ( \theta ) } < \min_{j \in \cS_{-1}^\theta}  \frac{Z_j ( \theta , x ) }{ P_{j , n} ( \theta ) } < \scrb$. In this case, we have
    \begin{equation}
    \begin{split}     
    &
        \int_\scra^\scrb f ( \theta , x_1, \ldots, x_n ) \, \d x_n 
    \\ & = \frac{ R ( \theta )}{i_n + 1 } \rbr*{ \textstyle \prod_{\ell=1}^{n-1} x_\ell^{i_\ell} } \br*{ \rbr*{ \min_{j \in \cS_{-1}^\theta}  \frac{Z_j ( \theta , x ) }{ P_{j , n} ( \theta ) }}^{i_n + 1} - \rbr*{ \max_{j \in \cS_1^\theta}  \frac{ Z_j ( \theta , x ) }{ P_{j , n} ( \theta ) }}^{i_n + 1} }.
    \end{split}
    \end{equation}
    \item It holds that $\scra < \max_{j \in \cS_1^\theta}  \frac{ Z_j ( \theta , x ) }{ P_{j , n} ( \theta ) } < \scrb \le \min_{j \in \cS_{-1}^\theta}  \frac{Z_j ( \theta , x ) }{ P_{j , n} ( \theta ) }$. In this case, we have
    \begin{equation}
        \int_\scra^\scrb f ( \theta , x_1, \ldots, x_n ) \, \d x_n = \frac{ R ( \theta )}{i_n + 1 } \rbr*{ \textstyle \prod_{\ell=1}^{n-1} x_\ell^{i_\ell} } \br*{ \scrb^{i_n + 1} - \rbr*{ \max_{j \in \cS_1^\theta}  \frac{ Z_j ( \theta , x ) }{ P_{j , n} ( \theta ) }}^{i_n + 1} }.
    \end{equation}
    \item It holds that $ \max_{j \in \cS_1^\theta}  \frac{ Z_j ( \theta , x ) }{ P_{j , n} ( \theta ) } \le \scra < \min_{j \in \cS_{-1}^\theta}  \frac{Z_j ( \theta , x ) }{ P_{j , n} ( \theta ) } < \scrb$. In this case, we have
    \begin{equation}
        \int_\scra^\scrb f ( \theta , x_1, \ldots, x_n ) \, \d x_n = \frac{ R ( \theta )}{i_n + 1 } \rbr*{ \textstyle \prod_{\ell=1}^{n-1} x_\ell^{i_\ell} } \br*{ \rbr*{ \min_{j \in \cS_{-1}^\theta}  \frac{Z_j ( \theta , x ) }{ P_{j , n} ( \theta ) }}^{i_n + 1} -\scra^{i_n + 1} }.
    \end{equation}
    \item It holds that $ \max_{j \in \cS_1^\theta}  \frac{ Z_j ( \theta , x ) }{ P_{j , n} ( \theta ) } \le \scra < \scrb \le \min_{j \in \cS_{-1}^\theta}  \frac{Z_j ( \theta , x ) }{ P_{j , n} ( \theta ) } $. In this case, we have
    \begin{equation}
        \int_\scra^\scrb f ( \theta , x_1, \ldots, x_n ) \, \d x_n = \frac{ R ( \theta )}{i_n + 1 } \rbr*{ \textstyle \prod_{\ell=1}^{n-1} x_\ell^{i_\ell} } \br*{\scrb^{i_n + 1} -\scra^{i_n + 1} }.
    \end{equation}
\end{itemize}
Since these four cases are disjoint, 
by summing over all possible choices of the sets $\cS^\theta_k$, $k \in \cu{-1, 0, 1}$, and all choices of subsets of $\cS^\theta_1$, $\cS_{-1}^\theta$ where the maximal/minimal values are achieved,
we can write
\begin{equation}
      \int_\scra^\scrb f ( \theta , x_1, \ldots, x_n ) \, \d x_n = \frac{ R ( \theta )}{i_n + 1 } \rbr*{ \textstyle \prod_{\ell=1}^{n-1} x_\ell^{i_\ell} } \br*{ (I) + (II) + (III) + (IV) },
\end{equation}
where
\begin{equation}
    \begin{split}
        (I) &= \sum_{ A \dot{\cup} B \dot{\cup} C = \cu{1, \ldots, r } } \br[\Bigg]{ \prod_{j \in A} \indicator{(0, \infty ) } ( P_{ j , n} ( \theta ) ) \prod_{j \in B} \indicator{(0, \infty ) } ( - P_{j , n} ( \theta ) ) \prod_{j \in C} \rbr*{ \indicator{ \cu{0 } } ( P_{j , n} ( \theta ) ) \indicator{[0, \infty ) } ( - Z_j ( \theta , x ) } } \\
        & \sum_{\emptyset \not= \cI \subseteq A} \sum_{\emptyset \not= \cJ \subseteq B} \Biggl[ \Biggl[
        \prod_{i \in \cI } \rbr*{ \indicator{(\scra , \scrb ) } \rbr*{ \frac{Z_i ( \theta , x ) }{ P_{i , n } ( \theta ) }} \indicator{\cu{0}} \rbr*{ \frac{ Z_i ( \theta , x )}{P_{i , n} ( \theta ) } - \frac{ Z_{\min \cI} ( \theta , x ) }{P_{ \min \cI , n} ( \theta ) } } } \Biggr. \Biggr. \\
        & \times \prod_{j \in A \backslash \cI } \indicator{(0, \infty ) } \rbr*{ \frac{ Z_{\min \cI} ( \theta , x ) }{P_{ \min \cI , n} ( \theta ) } - \frac{ Z_j ( \theta , x )}{P_{j , n} ( \theta ) } } 
        \prod_{i \in \cJ } \rbr*{ \indicator{(\scra , \scrb ) } \rbr*{ \frac{Z_i ( \theta , x ) }{ P_{i , n } ( \theta ) }} \indicator{\cu{0}} \rbr*{ \frac{ Z_i ( \theta , x )}{P_{i , n} ( \theta ) } - \frac{ Z_{\min \cJ} ( \theta , x ) }{P_{ \min \cJ , n} ( \theta ) } } } \\
        & \Biggl. \times \prod_{j \in B \backslash \cJ } \indicator{(0, \infty ) } \rbr*{ \frac{ Z_j ( \theta , x )}{P_{j , n} ( \theta ) } - \frac{ Z_{\min \cJ} ( \theta , x ) }{P_{ \min \cJ , n} ( \theta ) } } \indicator{(0, \infty ) } \rbr*{ \frac{ Z_{\min \cJ} ( \theta , x ) }{P_{ \min \cJ , n} ( \theta ) } - \frac{ Z_{\min \cI} ( \theta , x ) }{P_{ \min \cI , n} ( \theta ) } } \Biggr] \\
        & \Biggl. \times \br*{ \rbr*{ \frac{ Z_{\min \cJ} ( \theta , x ) }{P_{ \min \cJ , n } ( \theta ) } }^{i_n + 1 } - \rbr*{\frac{ Z_{\min \cI} ( \theta , x ) }{P_{ \min \cI , n} ( \theta ) } } ^{i_n + 1 } } \Biggr],
    \end{split}
\end{equation}
\begin{equation}
    \begin{split}
        (II) &= \sum_{ A \dot{\cup} B \dot{\cup} C = \cu{1, \ldots, r } } \br[\Bigg]{ \prod_{j \in A} \indicator{(0, \infty ) } ( P_{j , n} ( \theta ) ) \prod_{j \in B} \indicator{(0, \infty ) } ( - P_{j , n} ( \theta ) ) \prod_{j \in C} \rbr*{ \indicator{ \cu{0 } } ( P_{j , n} ( \theta ) ) \indicator{[0, \infty ) } ( - Z_j ( \theta , x ) } } \\
        & \sum_{\emptyset \not= \cI \subseteq A}  \Biggl[ \Biggl[
        \prod_{i \in \cI } \rbr*{ \indicator{(\scra , \scrb ) } \rbr*{ \frac{Z_i ( \theta , x ) }{ P_{ i , n } ( \theta ) }} \indicator{\cu{0}} \rbr*{ \frac{ Z_i ( \theta , x )}{P_{i , n} ( \theta ) } - \frac{ Z_{\min \cI} ( \theta , x ) }{P_{ \min \cI , n} ( \theta ) } } } \Biggr. \Biggr. \\
        & \times \prod_{j \in A \backslash \cI } \indicator{(0, \infty ) } \rbr*{ \frac{ Z_{\min \cI} ( \theta , x ) }{P_{ \min \cI , n} ( \theta ) } - \frac{ Z_j ( \theta , x )}{P_{j , n} ( \theta ) } } 
        \prod_{i \in B } \rbr*{ \indicator{[\scrb , \infty ) } \rbr*{ \frac{Z_i ( \theta , x ) }{ P_{i , n} ( \theta ) }} } \\
        & \Biggl. \times \br*{ \scrb^{i_n + 1 } - \rbr*{\frac{ Z_{\min \cI} ( \theta , x ) }{P_{ \min \cI , n} ( \theta ) } } ^{i_n + 1 } } \Biggr],
    \end{split}
\end{equation}
\begin{equation}
    \begin{split}
    (III) &=  \sum_{ A \dot{\cup} B \dot{\cup} C = \cu{1, \ldots, r } } \br[\Bigg]{ \prod_{j \in A} \indicator{(0, \infty ) } ( P_{j , n} ( \theta ) ) \prod_{j \in B} \indicator{(0, \infty ) } ( - P_{j , n} ( \theta ) ) \prod_{j \in C} \rbr*{ \indicator{ \cu{0 } } ( P_{j , n} ( \theta ) ) \indicator{[0, \infty ) } ( - Z_j ( \theta , x ) } } \\
        &  \sum_{\emptyset \not= \cJ \subseteq B} \Biggl[ \Biggl[
        \prod_{i \in A } \rbr*{ \indicator{(- \infty , \scra ] } \rbr*{ \frac{Z_i ( \theta , x ) }{ P_{ i , n } ( \theta ) }} } 
        \prod_{i \in \cJ } \rbr*{ \indicator{(\scra , \scrb ) } \rbr*{ \frac{Z_i ( \theta , x ) }{ P_{ i , n } ( \theta ) }} \indicator{\cu{0}} \rbr*{ \frac{ Z_i ( \theta , x )}{P_{ i , n} ( \theta ) } - \frac{ Z_{\min \cJ} ( \theta , x ) }{P_{ \min \cJ , n } ( \theta ) } } } \\
        & \Biggl. \times \prod_{j \in B \backslash \cJ } \indicator{(0, \infty ) } \rbr*{ \frac{ Z_j ( \theta , x )}{P_{j , n} ( \theta ) } - \frac{ Z_{\min \cJ} ( \theta , x ) }{P_{ \min \cJ , n} ( \theta ) } }  \Biggr] 
          \times \br*{ \rbr*{ \frac{ Z_{\min \cJ} ( \theta , x ) }{P_{ \min \cJ , n} ( \theta ) } }^{i_n + 1 } - \scra ^{i_n + 1 } } \Biggr],
    \end{split}
\end{equation}
and
\begin{equation}
    \begin{split}
         (IV) &=  \sum_{ A \dot{\cup} B \dot{\cup} C = \cu{1, \ldots, r } } \br[\Bigg]{ \prod_{j \in A} \indicator{(0, \infty ) } ( P_{j , n} ( \theta ) ) \prod_{j \in B} \indicator{(0, \infty ) } ( - P_{j , n} ( \theta ) ) \prod_{j \in C} \rbr*{ \indicator{ \cu{0 } } ( P_{j , n} ( \theta ) ) \indicator{[0, \infty ) } ( - Z_j ( \theta , x ) } } \\
        &  
       \times \rbr*{ \prod_{i \in A } \indicator{(- \infty , \scra ] } \rbr*{ \frac{Z_i ( \theta , x ) }{ P_{ i , n } ( \theta ) } }\prod_{i \in B } \indicator{[\scrb , \infty ) } \rbr*{ \frac{Z_i ( \theta , x ) }{ P_{i , n } ( \theta ) } }  }
         \br*{  \scrb^{i_n + 1 } - \scra ^{i_n + 1 } } .
    \end{split}
\end{equation}
Furthermore, \nobs that, e.g., in $(I)$ we have for all $i \in \cI \subseteq A$ that
\begin{equation}
    \begin{split}
        \indicator{(\scra , \scrb ) } \rbr*{ \frac{Z_i ( \theta , x ) }{ P_{ i , n } ( \theta ) }} &= \indicator{(\scra , \infty ) } \rbr*{ \frac{Z_i ( \theta , x ) }{ P_{ i , n } ( \theta ) }} \indicator{(- \infty , \scrb ) } \rbr*{ \frac{Z_i ( \theta , x ) }{ P_{i , n } ( \theta ) }} \\
        &= \indicator{(0, \infty ) } \rbr*{ Z_i ( \theta , x ) - \scra P_{ i , n} ( \theta ) } \indicator{(0, \infty ) } \rbr*{\scrb P_{ i , n} ( \theta ) - Z_i ( \theta , x ) }.
    \end{split}
\end{equation}
Similarly, the other indicator functions can be brought into the correct form, taking into account the different signs of $P_{j , n} ( \theta)$ for $j \in A$ and $j \in B$.
Moreover, \nobs that the remaining terms can be written as linear combinations of rational functions in $\theta$ and polynomials in $x$.
Hence, we obtain that the expressions $(I), (II), (III), (IV)$ are elements of $\scrA_{m, n-1}$.
\end{cproof}

\subsection{On the semialgebraic property of the risk function}
\label{subsection:risk:semialg}
\cfclear
\begin{definition} \label{def:multidim:piece:polyn}
Let $d \in \N$,
let $ A \subseteq \R^d$ be a set,
and let $f \colon A \to \R$ be a function. Then we say that $f$ is piecewise polynomial if and only if there exist $n \in \N$, 
$\alpha_1, \alpha_2, \ldots, \alpha_n \in \R^{n \times d}$, 
$\beta_{1}, \beta_2, \ldots, \beta_n \in \R^n$,
 $P_1, P_2, \ldots, P_n \in \polyn_d$ 
such that for all $x \in A$
it holds that
\begin{equation} \cfadd{def:polynomial}
        f(x) = \smallsum_{i=1}^n \br*{ P_i(x)  \indicator{[0, \infty )^n} \rbr{ \alpha_i x + \beta_i } } 
\end{equation}
\cfload.
\end{definition}

\cfclear
\begin{cor} \label{cor:loss:semialgebraic}
Assume \cref{setting:snn} and assume that $f$ and $\dens$ are piecewise polynomial \cfadd{def:multidim:piece:polyn}\cfload.
Then $\cL$ is semialgebraic \cfadd{def:semialgebraic:function}\cfload.
\end{cor}

\begin{cproof}{cor:loss:semialgebraic}
Throughout this proof let $F \colon \R^d \to \R$ and $\fP \colon \R^d \to \R$ satisfy for all $x \in \R^d$ that
\begin{equation} \label{cor:loss:semialgebraic:eqtilde}
    F ( x ) = \begin{cases} f(x) & \colon x \in [\scra , \scrb ] ^d \\
    0 & \colon x \notin [\scra , \scrb ] ^d
    \end{cases}
    \qqandqq
        \fP ( x ) = \begin{cases} \dens ( x ) & \colon x \in [\scra , \scrb ] ^d \\
    0 & \colon x \notin [\scra , \scrb ] ^d.
    \end{cases}
\end{equation}
\Nobs that \cref{cor:loss:semialgebraic:eqtilde} and the assumption that $f$ and $\dens$ are piecewise polynomial assure that 
\begin{equation} \label{cor:loss:semialgebraic:eq1} \cfadd{def:function:amn}
    \br*{  \R^\fd \times \R^d \ni (\theta , x ) \mapsto F ( x ) \in \R } \in \scrA_{\fd , d} \qandq \br*{  \R^\fd \times \R^d \ni (\theta , x ) \mapsto \fP ( x ) \in \R } \in \scrA_{\fd , d} 
\end{equation} 
\cfload.
In addition, \nobs that the fact that for all $\theta \in \R^\fd$, $x \in \R^d$ we have that
\begin{equation}
    \begin{split}
        \realization{\theta} ( x ) &= \c{\theta} +
        \sum_{i=1}^\width \v{\theta}_i \max \cu*{ \smallsum_{\ell =1}^d \w{\theta}_{i , \ell} x_\ell + \b{\theta}_i , 0 }  \\
        &=  \c{\theta} + \sum_{i=1}^\width \v{\theta}_i \rbr*{\smallsum_{\ell =1}^d \w{\theta}_{i , \ell} x_\ell + \b{\theta}_i} \indicator{[0, \infty ) } \rbr*{\smallsum_{\ell =1}^d \w{\theta}_{i , \ell} x_\ell + \b{\theta}_i}
    \end{split}
\end{equation}
demonstrates that
\begin{equation} \label{cor:loss:semialgebraic:eq:real:amn}
      \br*{  \R^\fd \times \R^d \ni (\theta , x ) \mapsto \realization{\theta} ( x ) \in \R } \in \scrA_{\fd , d} .
\end{equation}
Combining this with \cref{cor:loss:semialgebraic:eq1} and the fact that $\scrA_{\fd , d}$ is an algebra proves that
\begin{equation} 
    \br*{ \R^\fd \times \R^d \ni (\theta , x ) \mapsto ( \realization{\theta} ( x ) - F  ( x ) ) ^2 \fP ( x ) \in \R } \in \scrA_{\fd , d } .
\end{equation}
This,
\cref{prop:integrals:amn},
and induction demonstrate that
\begin{equation}
   \br*{ \R^\fd \ni \theta \mapsto  \int_\scra^\scrb \int_\scra^\scrb \cdots \int_\scra^\scrb ( \realization{\theta} ( x ) - F ( x ) ) ^2 \fP ( x ) \, \d x_d \cdots \, \d x_2 \, \d x_1 \in \R } \in \scrA_{\fd , 0}.
\end{equation}
Fubini's theorem hence implies that $\cL \in \scrA_{\fd , 0 }$.
Combining this and \cref{lem:amn:semialgebraic} shows that $\cL$ is semialgebraic.
\end{cproof}

\section{Convergence rates for solutions of GF differential equations}
\label{section:gf:loja}

In this section we employ the findings from \cref{section:risk:diff,section:semialgebraic} 
to establish in \cref{prop:gf:conv:local} in \cref{subsection:gf:local:conv} below, 
in \cref{prop:gf:convergence} in \cref{subsection:gf:local:conv}, 
and in \cref{theo:gf:conv:simple} in \cref{subsection:gf:global:conv} below 
several convergence rate results for solutions of GF differential equations. 
\cref{theo:intro:convergence} in the introduction is a direct consequence 
of \cref{theo:gf:conv:simple}. Our proof of \cref{theo:gf:conv:simple} 
is based on an application of \cref{prop:gf:convergence} and 
our proof of \cref{prop:gf:convergence} uses \cref{prop:gf:conv:local}. 
Our proof of \cref{prop:gf:conv:local}, in turn, employs \cref{prop:loss:lojasiewicz} 
in \cref{subsection:loja} below. In \cref{prop:loss:lojasiewicz} we establish that under the assumption that the target function $f \colon [ \scra , \scrb ] ^d \to \R$ and the unnormalized density function $\dens \colon [ \scra , \scrb ] ^d \to [0, \infty)$ are piecewise polynomial 
(see \cref{def:multidim:piece:polyn} in \cref{subsection:risk:semialg}) 
we have that the risk function $\cL \colon \R^\fd \to \R$ satisfies 
an appropriately generalized \L ojasiewicz inequality.

In the proof of \cref{prop:loss:lojasiewicz} the classical \L ojasiewicz inequality 
for semialgebraic or subanalytic functions (cf., e.g., Bierstone \& Milman~\cite{BierstoneMilman1988}) is 
not directly applicable since 
the generalized gradient function $ \cG \colon \R^{ \fd } \to \R^{ \fd } $ is not continuous. 
We will employ the more general results from Bolte et al.~\cite{BolteDaniilidis2006} 
which also apply to not necessarily continuously differentiable functions.

The arguments used in the proof of \cref{prop:gf:conv:local} are slight adaptions 
of well-known arguments in the literature; see, e.g.,
Kurdyka et al.~\cite[Section 1]{KurdykaMostowski2000},
Bolte et al.~\cite[Theorem 4.5]{BolteDaniilidis2006},
or Absil et al.~\cite[Theorem 2.2]{AbsilMahonyAndrews2005}. 
On the one hand, 
in Kurdyka et al.~\cite[Section 1]{KurdykaMostowski2000}
and Absil et al.~\cite[Theorem 2.2]{AbsilMahonyAndrews2005}
it is assumed that the object function of the considered optimization problem is analytic 
and 
in Bolte et al.~\cite[Theorem 4.5]{BolteDaniilidis2006} 
it is assumed that the objective function of the considered optimization problem 
is convex or lower $ C^2 $ 
and \cref{prop:gf:conv:local} does not require these assumptions. 
On the other hand, 
Bolte et al.~\cite[Theorem 4.5]{BolteDaniilidis2006} 
consider more general differential dynamics 
and the considered gradients are allowed to be more general than 
the specific generalized gradient function 
$ \cG \colon \R^{ \fd } \to \R^{ \fd } $ 
which is considered in \cref{prop:gf:conv:local}.

\subsection{Generalized \L ojasiewicz inequality for the risk function}
\label{subsection:loja}

\cfclear
\begin{prop}[Generalized {\L}ojasiewicz inequality] \label{prop:loss:lojasiewicz} 
Assume \cref{setting:snn}, 
assume that $\dens$ and $f$ are piecewise polynomial, and let $\vartheta \in \R^\fd$ \cfadd{def:multidim:piece:polyn}\cfload.
Then there exist $\varepsilon, \const \in (0, \infty)$, $\alpha \in ( 0 , 1 )$ such that for all $\theta \in B_\varepsilon ( \vartheta )$ it holds that
\begin{equation} \label{prop:loss:loja:eqclaim}
    \abs{\cL ( \theta ) - \cL ( \vartheta ) } ^\alpha \leq \const \norm {\cG ( \theta ) }.
\end{equation}
\end{prop}
\begin{cproof2}{prop:loss:lojasiewicz}
Throughout this proof let $\bfM \colon \R^\fd \to [0, \infty]$ satisfy for all $\theta \in \R^\fd$ that
\begin{equation}
    \bfM ( \theta ) = \inf \rbr*{ \cu*{\norm{h} \colon h \in \partial \cL ( \theta ) } \cup \cu{ \infty } }.
\end{equation}
\Nobs that \cref{prop:loss:gradient:subdiff} implies for all $\theta \in \R^\fd$ that $\bfM ( \theta ) \leq \norm{\cG ( \theta ) }$.
Furthermore, \nobs that
\cref{cor:loss:semialgebraic},
the fact that semialgebraic functions are subanalytic,
and Bolte et al.~\cite[Theorem 3.1 and Remark 3.2]{BolteDaniilidis2006} ensure that there exist $\varepsilon, \const \in (0, \infty)$, $\fa \in [ 0 , 1 )$ which satisfy for all $\theta \in B_\varepsilon ( \vartheta )$ that
\begin{equation} \label{prop:loss:loja:eq1}
    \abs{\cL ( \theta ) - \cL ( \vartheta ) } ^\fa \leq \const \bfM ( \theta ).
\end{equation}
Combining this with the fact that for all $\theta \in \R^\fd$ it holds that $\bfM ( \theta ) \leq \norm{\cG ( \theta ) }$ and the fact that $\sup_{\theta \in B_\varepsilon ( \vartheta ) } \abs{\cL ( \theta ) - \cL ( \vartheta ) } < \infty$ demonstrates that for all $\theta \in B_\varepsilon ( \vartheta )$, $\alpha \in (\fa , 1)$ we have that
\begin{equation}
\begin{split}
 \abs{\cL ( \theta ) - \cL ( \vartheta ) } ^{\alpha} 
 & \le  \abs{\cL ( \theta ) - \cL ( \vartheta ) } ^\fa \rbr[\big]{ \sup\nolimits_{\psi \in B_\varepsilon ( \vartheta ) } \abs{\cL ( \psi ) - \cL ( \vartheta ) } ^{\alpha - \fa} } \\
 &\le \rbr[\big]{\const \sup\nolimits_{\psi \in B_\varepsilon ( \vartheta ) } \abs{\cL ( \psi ) - \cL ( \vartheta ) } ^{\alpha - \fa} } \norm{\cG ( \theta ) }.
 \end{split}
\end{equation}
\end{cproof2}

\subsection{Local convergence for solutions of GF differential equations}
\label{subsection:gf:local:conv}

\begin{prop} \label{prop:gf:conv:local}
Assume \cref{setting:snn} and
let $\vartheta \in \R^\fd$, $\varepsilon, \const \in (0, \infty)$, $\alpha \in (0, 1)$ satisfy for all $\theta \in B_\varepsilon ( \vartheta )$ that
\begin{equation} \label{prop:gf:conv:local:eq:loj}
    \abs{\cL ( \theta ) - \cL ( \vartheta ) } ^\alpha \leq \const \norm{\cG ( \theta )}.
\end{equation}
Then there exists $\delta \in (0, \varepsilon) $ 
such that for all $\Theta \in C([0, \infty ), \R^\fd)$ with $\Theta_0 \in B_\delta ( \vartheta )$, 
$\forall \, t \in [0, \infty ) \colon \Theta_t = \Theta_0 - \int_0^t \cG ( \Theta_s ) \, \d s$, 
and $\inf_{t \in \cu{ s \in  [0, \infty ) \colon \Theta_s \in B_\varepsilon ( \vartheta ) } } \cL ( \Theta_t ) \geq \cL ( \vartheta )$
there exists $\psi \in \cL^{-1} (\cu{ \cL ( \vartheta ) } )$ such that 
for all $t \in [0 , \infty )$ it holds that 
$ \Theta_t \in B_\varepsilon ( \vartheta ) $, 
$ \int_0^\infty \norm{\cG ( \Theta_s ) } \, \d s \leq \varepsilon $, 
$ \abs{\cL ( \Theta_t ) - \cL ( \psi ) } \leq ( 1 + \const^{-2} t )^{-1} $,
and
\begin{equation} \label{prop:gf:conv:local:eq:statement}
  \norm{\Theta_t - \psi } \leq  \br*{ 1 + \rbr[\big]{ \const^{-\nicefrac{1}{\alpha}} ( 1 - \alpha )  }^{\frac{\alpha}{1 - \alpha } } t }^ { - \min \cu*{1, \frac{1 - \alpha}{ \alpha } } } .
\end{equation}
\end{prop}

\begin{cproof}{prop:gf:conv:local}
\Nobs that the fact that $\cL$ is continuous implies that there exists $\delta \in  (0, \nicefrac{\varepsilon }{ 3  } )$ which satisfies for all $\theta \in B_\delta ( \vartheta )$ that
\begin{equation} \label{prop:gf:convergence:eq:defdelta}
   \abs{\cL ( \theta ) - \cL ( \vartheta ) } ^{1 - \alpha }  \leq \min \cu*{ \frac{\varepsilon ( 1 - \alpha ) }{3 \const }, \frac{1 - \alpha}{\const} , 1 }.
\end{equation}
In the following let $\Theta \in C([0, \infty ) , \R^\fd)$ satisfy $\forall \, t \in [0, \infty ) \colon \Theta_t = \Theta_0 - \int_0^t \cG ( \Theta_s ) \, \d s$,
$\Theta_0 \in B_\delta ( \vartheta)$,
and 
\begin{equation} \label{prop:gf:convergence:eq:inf}
\inf\nolimits_{t \in \cu{ s \in  [0, \infty ) \colon \Theta_s \in B_\varepsilon ( \vartheta ) } } \cL ( \Theta_t ) \geq \cL ( \vartheta ).
\end{equation}
In the first step we show that for all $t \in [0, \infty )$ it holds that
\begin{equation} \label{prop:gf:convergence:eq:claim1}
\Theta_t \in B_\varepsilon ( \vartheta ).
\end{equation}
\Nobs that, e.g., \cite[Lemma 3.1]{JentzenRiekertFlow} ensures for all $t \in [0, \infty)$ that
\begin{equation} \label{prop:gf:convergence:eq:lossintegral}
    \cL ( \Theta_t ) = \cL ( \Theta_0 ) - \int_0^t \norm{\cG ( \Theta_s ) } ^2 \, \d s.
\end{equation}
This implies that $[0, \infty) \ni t \mapsto \cL ( \Theta_t ) \in [0, \infty)$ is non-increasing.
Next let $L \colon [0, \infty ) \to \R$ satisfy for all $t \in [0, \infty)$ that 
\begin{equation} \label{prop:gf:convergence:eq:defl}
L(t) = \cL ( \Theta_t ) - \cL ( \vartheta )    
\end{equation}
and let $T \in [0 , \infty]$ satisfy
\begin{equation} \label{prop:gf:convergence:eq:tout}
    T = \inf \rbr*{ \cu*{t \in [0, \infty ) \colon \norm{\Theta_t - \vartheta } \geq \varepsilon  } \cup \cu{\infty} }.
\end{equation}
We intend to show that $T = \infty$. 
\Nobs that \cref{prop:gf:convergence:eq:inf} assures for all $t \in [0, T)$ that $L(t) \geq 0$. 
Moreover, \nobs that \cref{prop:gf:convergence:eq:lossintegral} and \cref{prop:gf:convergence:eq:defl} ensure that for almost all $t \in [ 0 , T)$ it holds that $L$ is differentiable at $t$ and satisfies $ L ' ( t ) = \frac{\d}{\d t} ( \cL ( \Theta_t ) ) = - \norm{\cG ( \Theta_t ) } ^2$.
In the following let $\tau \in [0, T]$ satisfy
\begin{equation}
    \tau = \inf \rbr*{ \cu*{t \in [0, T) \colon L ( t ) = 0 } \cup \cu{T } }.
\end{equation}
 \Nobs that the fact that $L$ is non-increasing
implies that for all $s \in [ \tau , T)$ it holds that $L(s) = 0$. 
Combining this with \cref{prop:gf:convergence:eq:lossintegral} demonstrates for almost all $s \in (\tau, T)$ that $\cG ( \Theta_s ) = 0$. This proves for all $s \in [\tau, T)$ that $\Theta_s = \Theta_\tau$.
Next \nobs that \cref{prop:gf:conv:local:eq:loj} ensures that for all $t \in [0, \tau)$ it holds that
\begin{equation}
    0 < [ L ( t ) ] ^\alpha = \abs{\cL ( \Theta_t ) - \cL ( \vartheta ) } ^\alpha \leq \const \norm{\cG ( \Theta_t ) }.
\end{equation}
Combining this with 
the chain rule
proves for almost all $t \in [0, \tau)$ that
\begin{equation} \label{prop:gf:conv:eq:derivative}
\begin{split}
    \frac{\d }{\d t} ([ L ( t ) ]^{1 - \alpha } ) &= (1-\alpha) [ L ( t ) ]^{-\alpha} \rbr*{ - \norm{\cG(\Theta_t)}^2  } 
    \\
&\leq - ( 1 - \alpha ) \const^{-1} \norm{\cG ( \Theta_t ) }^{-1} \norm{\cG ( \Theta_t ) } ^2 = - \const^{-1} (1 - \alpha ) \norm{\cG(\Theta_t)}.
\end{split}
\end{equation}
In addition, \nobs that the fact that $[0, \infty ) \ni t \mapsto L(t) \in \R$ is absolutely continuous and the fact that for all $r \in (0, \infty)$ it holds that $r , \infty ) \ni y \mapsto y^{1 - \alpha } \in \R$ is Lipschitz continuous demonstrate for all $t \in [0, \tau )$ that $[0, t] \ni s \mapsto [ L ( s ) ]^{ 1 - \alpha } \in \R$ is absolutely continuous.
Integrating \cref{prop:gf:conv:eq:derivative} hence shows for all $s , t \in [0, \tau )$ with $t \le s$
that
\begin{equation} 
\int _t ^s \norm{\cG(\Theta_u ) } \, \d u \leq - \const \rbr{1 - \alpha } ^{-1} \rbr {  [ L ( s ) ] ^{1 - \alpha } - [ L ( t ) ] ^{ 1 - \alpha }  } \leq \const \rbr{1 - \alpha}^{-1} [ L ( t ) ]^{1 - \alpha} .
\end{equation}
This and the fact that for almost all $s \in (\tau, T)$ it holds that $\cG ( \Theta_s ) = 0$ ensure that for all $s , t \in [0, T)$ with $t \le s$ we have that
\begin{equation} \label{prop:gf_conv:eq:integrated}
    \int_t^s \norm{\cG(\Theta_u ) } \, \d u 
    \leq \const \rbr{ 1 - \alpha}^{-1} [ L ( t ) ]^{1 - \alpha} .
\end{equation}
Combining this with \cref{prop:gf:convergence:eq:defdelta} demonstrates for all $t \in [0, T)$
that
\begin{equation} \label{prop:gf:convergence:eq:tailbound}
\norm{\Theta_t - \Theta_{0} } = \norm*{\int_{0}^t \cG ( \Theta_s )  \, \d s} \leq \int _{0} ^t \norm{\cG(\Theta_s ) } \, \d s \leq \frac{\const \abs{\cL ( \Theta_0 ) - \cL ( \vartheta ) }^{1-\alpha} }{1 - \alpha} \leq \min \cu*{ \frac{\varepsilon}{3} , 1 }.
\end{equation}
This, the fact that $\delta < \nicefrac{\varepsilon}{3}$, and the triangle inequality assure for all $t \in [0, T)$ that
\begin{equation} 
\norm{\Theta_t - \vartheta } \leq  \norm{\Theta_t - \Theta_{0} } + \norm{\Theta_{0} - \vartheta}  \leq  \frac{\varepsilon}{3} + \delta \leq \frac{\varepsilon}{3} + \frac{\varepsilon}{3} = \frac{2 \varepsilon }{3 }.
\end{equation}
Combining this with \cref{prop:gf:convergence:eq:tout} proves that $T = \infty$. This establishes \cref{prop:gf:convergence:eq:claim1}.

Next \nobs that the fact that $T = \infty$ and \eqref{prop:gf:convergence:eq:tailbound} prove that 
\begin{equation} \label{prop:gf:convergence:eq:trajectory}
 \int_0^\infty \norm{\cG ( \Theta_s ) } \, \d s 
\leq  \min \cu*{ \frac{\varepsilon}{3} , 1 } \le \varepsilon < \infty.
\end{equation}
In the following let $\sigma \colon [0, \infty ) \to [ 0 , \infty)$ satisfy for all $t \in [0, \infty)$ that
\begin{equation}
    \sigma ( t ) = \int_t^\infty \norm{\cG ( \Theta_s )} \, \d s.
\end{equation}
\Nobs that \cref{prop:gf:convergence:eq:trajectory} proves that $\limsup_{t \to \infty} \sigma ( t ) = 0$. 
In addition, \nobs that \cref{prop:gf:convergence:eq:trajectory} assures that there exists $\psi \in\R^\fd$ such that
\begin{equation} \label{prop:gf:convergence:eq:limitpsi}
    \limsup \nolimits_{t \to \infty} \norm{\Theta_t - \psi } = 0.
\end{equation}
In the next step we combine the weak chain rule for the risk function in \cref{prop:gf:convergence:eq:lossintegral} with \cref{prop:gf:convergence:eq:claim1} and \cref{prop:gf:conv:local:eq:loj} to obtain that for almost all $t \in [0, \infty)$ we have that
\begin{equation}
L ' ( t )  = - \norm{\cG ( \Theta_t ) } ^2 \leq - \const^{-2} [ L ( t ) ] ^{ 2 \alpha}.
\end{equation}
In addition, \nobs that the fact that $L$ is non-increasing and \cref{prop:gf:convergence:eq:defdelta} ensure that for all $t \in [0, \infty)$ it holds that $L ( t ) \leq L ( 0 ) \leq 1$.
Therefore, we get for almost all $t \in [0, \infty)$ that
\begin{equation}
   L ' ( t )
    \leq - \const^{-2} [ L ( t ) ] ^{ 2 }.
\end{equation}
Combining this with the fact that for all $t \in [0, \tau)$ it holds that $L ( t ) > 0$
establishes for almost all $t \in [0, \tau)$ that
\begin{equation}
    \frac{\d}{\d t} \rbr*{ \frac{\const^2}{L ( t ) } } = - \frac{\const^2 L ' ( t )}{[ L ( t ) ] ^2} \geq 1.
\end{equation}
The fact that for all $t \in [0, \tau )$ it holds that $[0, t ] \ni s \mapsto L ( s ) \in (0, \infty)$ is absolutely continuous hence demonstrates for all $t \in [0, \tau)$ that
\begin{equation}
    \frac{\const^2}{L(t)} \geq \frac{\const^2}{L ( 0 )} + t \geq \const^2 + t.
\end{equation}
Therefore, we infer for all $t \in [0, \tau)$ that
\begin{equation}
    L ( t ) \leq \const^2 \rbr*{ \const^2 + t }^{-1} = \rbr*{ 1 + \const^{-2}t }^{-1}.
\end{equation}
This and the fact that for all $t \in [\tau, \infty)$ it holds that $L(t) = 0$ prove that for all $t \in [0, \infty )$ we have that 
\begin{equation} \label{prop:gf:convergence:eq:lossest}
\abs{\cL ( \Theta_t ) - \cL ( \vartheta ) } = L ( t ) \leq \rbr*{ 1 + \const^{-2}t }^{-1}.
\end{equation}
 Furthermore, \nobs that \cref{prop:gf:convergence:eq:limitpsi} and the fact that $\cL$ is continuous imply that $\limsup_{t \to \infty} \abs{\cL ( \Theta_t ) - \cL ( \psi ) } = 0$. Hence, we obtain that $\cL ( \psi ) = \cL ( \vartheta )$. This shows for all $t \in [0, \infty )$ that
\begin{equation} \label{prop:gf:convergence:eq:claim2}
   \abs{\cL ( \Theta_t ) - \cL ( \psi ) } \leq \rbr*{ 1 + \const^{-2}t }^{-1}.
\end{equation}
In the next step we establish a convergence rate for the quantity $\norm{\Theta_t - \psi}$, $t \in [0, \infty )$.
We accomplish this by employing an upper bound for the tail length of the curve $\Theta_t \in \R^\fd$, $t \in [0, \infty )$.
More formally,
\nobs that \cref{prop:gf_conv:eq:integrated},
\cref{prop:gf:convergence:eq:claim1},
and \cref{prop:gf:conv:local:eq:loj} demonstrate for all $t \in [0, \infty )$ that
\begin{equation} \label{prop:gf:conv:eq:tailest}
\begin{split}
\sigma(t) 
&= \int_t^\infty \norm{\cG ( \Theta_u )} \, \d u = \lim_{s \to \infty} \br*{ \int_t^s \norm{\cG ( \Theta_u ) } \, \d u } \\
& \leq \const \rbr{1 - \alpha}^{-1}  \br{L ( t ) }^{1-\alpha}
\leq \const \rbr{1-\alpha}^{-1} \rbr*{ \const  \norm{\cG(\Theta_t) }  }^{\frac{1-\alpha}{\alpha}}.
\end{split}
\end{equation}
Next \nobs that the fact that for all $t \in [0, \infty )$ it holds that $\sigma ( t ) = \int_0^\infty \norm{\cG ( \Theta_s )} \, \d s - \int_0^t \norm{\cG ( \Theta_s ) } \, \d s$ shows that for almost all $t \in [ 0 , \infty )$ we have that $\sigma' ( t ) = - \norm{\cG ( \Theta_t ) }$. This and \cref{prop:gf:conv:eq:tailest} yield for almost all $t \in [0, \infty )$ that $\sigma(t) \leq \const^{\nicefrac{1}{\alpha}} \rbr{1-\alpha}^{-1} \br*{ - \sigma ' (t)   }^{\frac{1-\alpha}{\alpha}}$.
Therefore, we obtain for almost all $t \in [0, \infty )$ that
\begin{equation}
\sigma ' ( t ) \leq - \br[\big]{  ( 1-\alpha ) \const^{-\nicefrac{1}{\alpha}} \sigma ( t )  }^{\frac{\alpha}{1-\alpha}}.
\end{equation}
Combining this with the fact that $\sigma$ is absolutely continuous implies for all $t \in [0, \infty )$ that
\begin{equation} \label{prop:gf:convergence:eq:sigmaint}
\sigma(t) - \sigma(0) \leq - \br[\big]{  ( 1-\alpha ) \const^{-\nicefrac{1}{\alpha}} }^{\frac{\alpha}{1-\alpha}} \int _{0}^t [ \sigma ( s ) ]^{\frac{\alpha}{1-\alpha}} \,  \d s.
\end{equation}
In the following let $\beta, \fC \in (0, \infty)$ satisfy $\beta = \max \cu{ 1 , \frac{\alpha}{1-\alpha} }$ and $\fC =\rbr*{ ( 1-\alpha ) \const ^{ -\nicefrac{1}{\alpha}}  }^{\frac{\alpha}{1-\alpha}}$. 
\Nobs that \cref{prop:gf:convergence:eq:sigmaint} and the fact that for all $t \in [0, \infty)$ it holds that $\sigma ( t ) \leq \sigma ( 0 ) \leq 1$ ensure that for all $t \in [0, \infty)$ it holds that
\begin{equation}
\sigma(t) \leq \sigma(0) - \fC \int _{0}^t [ \sigma ( s ) ]^\beta \, \d s.
\end{equation}
This, the fact that $\sigma$ is non-increasing,
and the fact that for all $t \in [0, \infty )$ it holds that $0 \leq \sigma(t) \leq 1$ prove that for all $t \in [0, \infty )$ we have that
\begin{equation}
\br{ \sigma ( t ) }^\beta \le \sigma ( t ) \leq \sigma(0) -\fC [ \sigma ( t ) ]^\beta t \leq 1 - \fC t [ \sigma ( t ) ]^\beta .
\end{equation}
Hence, we obtain for all $t \in [0, \infty )$ that $\sigma(t) \leq \rbr*{ 1 + \fC t }^{-\frac{1}{\beta}}$.
Combining this with the fact that for all $t \in [0, \infty)$ it holds that
\begin{equation} 
\begin{split}
    \norm{\Theta_t - \psi} &\leq \limsup_{s \to \infty} \norm{\Theta_t - \Theta_s } = \limsup_{s \to \infty} \norm*{ \int_t^s \cG ( \Theta_u )  \, \d u } \leq \limsup_{s \to \infty} \br*{ \int_t^s \norm{\cG ( \Theta_u ) } \, \d u } \\
    &= \int_t^\infty \norm{\cG ( \Theta_u ) } \, \d u = \sigma ( t )
\end{split}
\end{equation}
shows that for all $t \in [0, \infty )$ we have that $\norm{\Theta_t - \psi } \le ( 1 + \fC t ) ^{- \nicefrac{1}{\beta}}$.
This, \cref{prop:gf:convergence:eq:claim1}, \cref{prop:gf:convergence:eq:trajectory}, 
and \cref{prop:gf:convergence:eq:claim2} establish \cref{prop:gf:conv:local:eq:statement}.
\end{cproof}

\subsection{Global convergence for solutions of GF differential equations}
\label{subsection:gf:global:conv}

\cfclear
\begin{prop} \label{prop:gf:convergence}
Assume \cref{setting:snn},
assume that $\dens$ and $f$ are piecewise
polynomial, 
and let $\Theta \in C ( [ 0 , \infty ) , \R^\fd )$ satisfy $\liminf_{t \to  \infty } \norm{\Theta_t } < \infty$ and $\forall \, t \in [0, \infty ) \colon \Theta_t = \Theta_0 - \int_0^t \cG ( \Theta_s ) \, \d s$ \cfadd{def:multidim:piece:polyn}\cfload.
Then there exist $\vartheta \in \cG^{ - 1 } ( \cu{  0 } )$,
$ \fC, \tau , \beta \in (0, \infty)$ which satisfy for all $t \in [ \tau , \infty )$ that
\begin{equation} \label{theo:gf:convergence:eq1}
    \norm{\Theta_t - \vartheta} \leq  \rbr[\big]{ 1 + \fC ( t - \tau ) }^{- \beta} 
\qqandqq
  \abs{ \cL ( \Theta_t ) - \cL ( \vartheta ) } \leq  \rbr[\big]{1 + \fC ( t - \tau ) } ^{-1}.
\end{equation}
\end{prop}

\begin{cproof}{prop:gf:convergence}
First \nobs that \cite[Lemma 3.1]{JentzenRiekertFlow} ensures that for all $t \in [0, \infty)$ it holds that
\begin{equation} \label{proof:gf:convergence:eq:lossintegral}
    \cL ( \Theta_t ) = \cL ( \Theta_0 ) - \int_0^t \norm{\cG ( \Theta_s ) } ^2 \, \d s.
\end{equation}
This implies that $[0, \infty) \ni t \mapsto \cL ( \Theta_t ) \in [0, \infty)$ is non-increasing. Hence,
we obtain that
 there exists $\bfm \in [0, \infty)$ which satisfies that
\begin{equation} \label{proof:gf:convergence:eq:defm}
    \bfm = \limsup\nolimits_{t \to \infty} \cL ( \Theta_t ) = \liminf\nolimits_{t \to \infty} \cL ( \Theta_t ) = \inf\nolimits_{t \in [0, \infty )} \cL ( \Theta_t ).
\end{equation}
Moreover,
\nobs that the assumption that $\liminf_{t \to  \infty } \norm{\Theta_t } < \infty$ ensures that there exist $\vartheta \in \R^\fd$ and $\tau = (\tau_n)_{n \in \N} \colon \N \to [0, \infty)$ which satisfy $\liminf_{n \to \infty} \tau_n = \infty$ and 
\begin{equation} \label{proof:gf:convergence:eq:taun}
	\limsup\nolimits_{n \to \infty} \norm{\Theta_{\tau_n} - \vartheta } = 0.
\end{equation} 
Combining this with \cref{proof:gf:convergence:eq:defm} and the fact that $\cL$ is continuous shows that 
\begin{equation} \label{proof:gf:convergence:eq:bfm}
	\cL ( \vartheta ) = \bfm \qqandqq \forall \, t \in [0, \infty ) \colon \cL ( \Theta_t ) \geq \cL ( \vartheta ).
\end{equation}
Next \nobs that \cref{prop:loss:lojasiewicz} demonstrates that there exist $\varepsilon, \const \in (0, \infty)$,
$\alpha \in ( 0 , 1 )$ such that for all $\theta \in B_\varepsilon ( \vartheta )$ we have that 
\begin{equation} \label{proof:gf:convergence:eq:loj}
    \abs{\cL ( \theta ) - \cL ( \vartheta ) }^\alpha \leq \const \norm{ \cG ( \theta ) } .
\end{equation}
Combining this and \cref{proof:gf:convergence:eq:taun} with \cref{prop:gf:conv:local} proves 
that there exists $\delta \in (0, \varepsilon)$ which satisfies for all $\Phi \in C([0, \infty ) , \R^\fd)$ with
 $\Phi_0 \in B_\delta ( \vartheta )$,
  $\forall \, t \in [0, \infty ) \colon \Phi_t = \Phi_0 - \int_0^t \cG ( \Phi_s ) \, \d s$,
   and
$\inf_{t \in \cu{s \in [0, \infty ) \colon \Phi_s \in B_\varepsilon ( \vartheta ) } } \cL ( \Phi_t ) \ge \cL ( \vartheta)$
 that it holds for all $t \in [0, \infty )$ that
 $\Phi_t \in B_\varepsilon ( \vartheta )$,
$\abs{\cL ( \Phi_t ) - \cL ( \vartheta ) } \leq ( 1 + \const^{-2} t )^{-1}$,
and
\begin{equation} \label{proof:gf:convergence:eq:convspeed}
\norm{\Phi_t - \vartheta } \leq  \br*{ 1 + \rbr[\big]{ \const^{-\nicefrac{1}{\alpha}} ( 1 - \alpha )  }^{\frac{\alpha}{1 - \alpha } } t }^ { - \min \cu*{1, \frac{1 - \alpha}{ \alpha } } } .
\end{equation}
Moreover, \nobs that \cref{proof:gf:convergence:eq:taun}  ensures that there exists $n \in \N$ which satisfies $\Theta_{\tau_n } \in B_\delta ( \vartheta )$.
Next let $\Phi \in C([0, \infty ) , \R^\fd)$ satisfy for all $t \in [0, \infty )$ that
\begin{equation} \label{proof:gf:convergence:eq:defphi}
\Phi_t = \Theta_{t + \tau_n }.
\end{equation}
\Nobs that \cref{proof:gf:convergence:eq:defphi,proof:gf:convergence:eq:bfm} assure that 
\begin{equation}
\Phi_0 \in B_\delta ( \vartheta ) , \quad \inf\nolimits_{t \in [0, \infty ) } \cL ( \Phi_t ) \ge \cL ( \vartheta ) , \qandq \forall \, t \in [0, \infty ) \colon \Phi_t = \Phi_0 - \int_0^t \cG ( \Phi_s ) \, \d s.
\end{equation}
Combining this with \cref{proof:gf:convergence:eq:convspeed} proves for all $t \in [\tau_n , \infty )$ that
\begin{equation} \label{proof:gf:convergence:eq:conc1}
\abs{\cL ( \Theta_t ) - \cL ( \vartheta ) } \le \rbr*{ 1 + \const^{-2} ( t - \tau_n ) }^{-1} 
\end{equation}
and
\begin{equation} \label{proof:gf:convergence:eq:conc2}
 \norm{\Theta_t - \vartheta } \leq  \br*{ 1 + \rbr[\big]{ \const^{-\nicefrac{1}{\alpha}} ( 1 - \alpha )  }^{\frac{\alpha}{1 - \alpha } } ( t - \tau_n ) }^ { - \min \cu*{1, \frac{1 - \alpha}{ \alpha } } }  .
\end{equation}
Next \nobs that \cite[Corollary 2.16]{JentzenRiekertFlow} shows that $\R^\fd \ni \theta \mapsto \norm{\cG ( \theta) } \in [0, \infty )$ is lower semicontinuous.
The fact that $\liminf_{s \to \infty} \norm{\cG ( \Theta_s ) } = 0$ and the fact that $\limsup_{t \to \infty} \norm{\Theta_t - \vartheta } = 0$ hence imply that $\cG ( \vartheta ) = 0$.
Combining this with \cref{proof:gf:convergence:eq:conc1} and \cref{proof:gf:convergence:eq:conc2} establishes \cref{theo:gf:convergence:eq1}.
\end{cproof}

\cfclear
\begin{theorem} \label{theo:gf:conv:simple}
Assume \cref{setting:snn},
assume that $\dens$ and $f$ are piecewise
polynomial, 
and let $\Theta \in C ( [ 0 , \infty ) , \R^\fd )$ satisfy $\liminf_{t \to  \infty } \norm{\Theta_t } < \infty$ and $\forall \, t \in [0, \infty ) \colon \Theta_t = \Theta_0 - \int_0^t \cG ( \Theta_s ) \, \d s$ \cfadd{def:multidim:piece:polyn}\cfload.
Then there exist $\vartheta \in \cG^{ - 1 } ( \cu{  0 } )$,
$\scrC,  \beta \in (0, \infty)$ which satisfy for all $t \in [ 0 , \infty )$ that
\begin{equation} \label{cor:gf:convergence:eq1}
    \norm{\Theta_t - \vartheta} \leq  \scrC ( 1 + t ) ^{ - \beta }
\qqandqq
  \abs{ \cL ( \Theta_t ) - \cL ( \vartheta ) } \leq  \scrC ( 1 + t ) ^{ - 1 } .
\end{equation}
\end{theorem}
\begin{cproof2}{theo:gf:conv:simple}
\Nobs that \cref{prop:gf:convergence} assures that
 there exist $\vartheta \in \cG^{ - 1 } ( \cu{  0 } )$,
$ \fC, \tau , \beta \in (0, \infty)$ which satisfy for all $t \in [ \tau , \infty )$ that
\begin{equation} \label{proof:cor:gf:convergence:eq1}
    \norm{\Theta_t - \vartheta} \leq  \rbr[\big]{ 1 + \fC ( t - \tau ) }^{- \beta} 
\end{equation}
and
\begin{equation} \label{proof:cor:gf:convergence:eq2}
  \abs{ \cL ( \Theta_t ) - \cL ( \vartheta ) } \leq  \rbr[\big]{1 + \fC ( t - \tau ) } ^{-1}.
\end{equation}
In the following let $\scrC \in (0, \infty)$ satisfy
\begin{equation} \label{proof:cor:gf:conv:eq3}
    \scrC = \max \cu*{\fC^{-1} , 1 + \tau , \fC^{- \beta}, (1 + \tau )^\beta , ( 1 + \tau ) ^\beta \br[\big]{ \sup\nolimits_{s \in [0, \tau ] } \norm{\Theta_s - \vartheta}  } , (1 + \tau ) \cL ( \Theta_0 )  }.
\end{equation}
\Nobs that
\cref{proof:cor:gf:convergence:eq2}, \cref{proof:cor:gf:conv:eq3},
 and the fact that $[0, \infty ) \ni t \mapsto \cL ( \Theta_t ) \in [0, \infty )$ is non-increasing show for all $t \in [0, \tau ]$ that
\begin{equation}
    \norm{\Theta_t - \vartheta }  \le \sup\nolimits_{s \in [0,  \tau ]} \norm{\Theta_s - \vartheta} \le \scrC ( 1 + \tau ) ^{- \beta } \le \scrC ( 1 + t ) ^{-\beta }
\end{equation}
and
\begin{equation}
    \abs{\cL ( \Theta_t ) - \cL ( \vartheta ) } = \cL ( \Theta_t ) - \cL ( \vartheta) 
    \le \cL ( \Theta_t ) \le  \cL ( \Theta_0 ) \le \scrC ( 1 + \tau )^{-1} \le \scrC ( 1 + t ) ^{-1}. 
\end{equation}
Moreover, \nobs that \cref{proof:cor:gf:convergence:eq1,proof:cor:gf:conv:eq3} imply for all $t \in [ \tau , \infty ) $ that
\begin{equation}
        \norm{\Theta_t - \vartheta } \leq  \scrC  \rbr[\big]{ \scrC^{\nicefrac{1}{\beta}} + \fC \scrC^{\nicefrac{1}{\beta}} ( t - \tau ) } ^{ - \beta } 
         \le \scrC \rbr[\big]{ \scrC^{\nicefrac{1}{\beta}} - \tau + t  }^{-\beta} \le \scrC (1 +  t ) ^{-\beta } .
\end{equation}
In addition, \nobs that \cref{proof:cor:gf:convergence:eq2,proof:cor:gf:conv:eq3} demonstrate for all $t \in [ \tau , \infty ) $ that
\begin{equation}
        \abs{\cL ( \Theta_t ) - \cL ( \vartheta ) } 
        \le \scrC \rbr[\big]{ \scrC + \fC \scrC ( t - \tau ) } ^{-1}
         \le \scrC \rbr[\big]{ \scrC - \tau + t }^{-1} \le \scrC ( 1 + t ) ^{-1} .
\end{equation}
\end{cproof2}

\subsection*{Acknowledgements}
This work has been funded by the Deutsche Forschungsgemeinschaft
(DFG, German Research Foundation) under Germany’s Excellence Strategy EXC 2044-390685587, Mathematics Münster: Dynamics-Geometry-Structure.


\end{document}